\documentclass[a4paper,10pt,twocolumn,preprint,5p]{elsarticle}
\usepackage{graphicx,color}
\usepackage[utf8]{inputenc}
\usepackage[T1]{fontenc}
\usepackage{bbm}
\usepackage{color}
\usepackage{amssymb}
\usepackage{multirow}
\usepackage{lineno}
\usepackage{hyperref}
\usepackage{algorithm}
\usepackage[noend]{algpseudocode}
\usepackage{amsmath}
\usepackage{subfig}
\usepackage{amsthm}
\modulolinenumbers[5]

\journal{Journal of Expert Systems with applications}
\newcommand\Tstrut{\rule{0pt}{2.6ex}}         
\newcommand\Bstrut{\rule[-0.9ex]{0pt}{0pt}}   

\biboptions{authoryear}

\newtheorem{theorem}{Theorem}[section]

\newtheorem{lemma}[theorem]{Lemma}

\usepackage{lipsum}
\makeatletter
\def\ps@pprintTitle{%
 \let\@oddhead\@empty
 \let\@evenhead\@empty
 \def\@oddfoot{http://dx.doi.org/10.1016/j.eswa.2016.10.012}%
 \let\@evenfoot\@oddfoot}
\makeatother

\begin{document}

\begin{frontmatter}

\title{A framework for redescription set construction}

\author[myfootnote]{Matej Mihelčić \corref{cor1}\fnref{mps}}
\ead{matej.mihelcic@irb.hr}
\cortext[cor1]{Corresponding author. Tel. +385 (1) 456 1080}
\author[myfootnote1]{Sašo Džeroski\fnref{mps}}
\ead{saso.dzeroski@ijs.si}
\author[myfootnote1]{Nada Lavrač\fnref{mps}}
\ead{nada.lavrac@ijs.si}

\address[myfootnote]{Ruđer Bošković Institute, Bijenička cesta 54, 10000 Zagreb, Croatia}
\address[myfootnote1]{Jožef Stefan Institute, Jamova cesta 39, 1000 Ljubljana, Slovenia}
\address[mps]{International Postgraduate School, Jamova cesta 39, 1000 Ljubljana, Slovenia}

\author{Tomislav Šmuc\fnref{myfootnote}}
\ead{tomislav.smuc@irb.hr}

\begin{abstract}
 
Redescription mining is a field of knowledge discovery that aims at finding different descriptions of similar subsets of instances in the data. These descriptions are represented as rules inferred from one or more disjoint sets of attributes, called views. As such, they support knowledge discovery process and help domain experts in formulating new hypotheses or constructing new knowledge bases and decision support systems. In contrast to previous approaches that typically create one smaller set of  redescriptions satisfying  a pre-defined set of constraints, we introduce a framework that creates large and heterogeneous redescription set from which user/expert can extract compact sets of differing properties,  according to its own preferences. Construction of large and heterogeneous redescription set relies on CLUS-RM algorithm and a novel, conjunctive refinement procedure that facilitates generation of larger and more accurate redescription sets. The work also introduces the variability of redescription accuracy when missing values are present in the data, which significantly extends applicability of the method.  Crucial part of the framework is the redescription set extraction based on heuristic multi-objective optimization procedure that allows user to define importance levels towards one or more redescription quality criteria.  We provide both theoretical and empirical comparison of the novel framework against current state of the art redescription mining algorithms and show that it represents more efficient and versatile approach for mining redescriptions from data.

\end{abstract}


\end{frontmatter}

\section{Introduction}

\noindent  In many scientific fields, there is a growing need to understand measured or observed data, to find different regularities or anomalies, groups of instances (patterns) for which they occur and their descriptions in order to get an insight into the underlying phenomena. 

This is addressed by redescription mining \citep{RamakrishnanCart}, a type of knowledge discovery that aims to find different descriptions of similar sets of instances by using one, or more disjoint sets of descriptive attributes, called views. It is applicable in a variety of scientific fields like  biology, economy, pharmacy, ecology, social science and other, where it is important to understand connections between different descriptors and to find regularities that are valid for different subsets of instances. Redescriptions are tuples of logical formulas which are called queries. Redescription $R_{ex}=(q_1,q_2)$ contains two queries: 

\noindent $q_1:\ (-1.8 \leq \tilde{t}_{7} \leq 4.4 \wedge 12.1 \leq \tilde{p}_{6} \leq 21.2)$\par \noindent
$q_2:\ \text{Polarbear}$\par \noindent

\noindent The first query ($q_1'$) describes a set of instances (geospatial locations) by using a set of attributes related to temperature ($t$) and precipitation ($p$) in a given month as first view (in the example average temperature in July and average precipitation in June). The second query ($q_2'$) describes very similar set of locations by using a set of attributes specifying animal species inhabiting these locations as a second view (in this instance polar bear). Queries contain only conjunction logical operator, though the approach supports conjunction, negation and disjunction operators.

 We first describe the fields of data mining and knowledge discovery closely related to redescription mining. Next, we describe recent research in redescription mining, relevant to the approach we propose. We then outline our approach positioned in the context of related work. 

\subsection{Fields related to redescription mining}
\begin{figure*}[ht]
    \centering
    \includegraphics[width=0.62\textwidth]{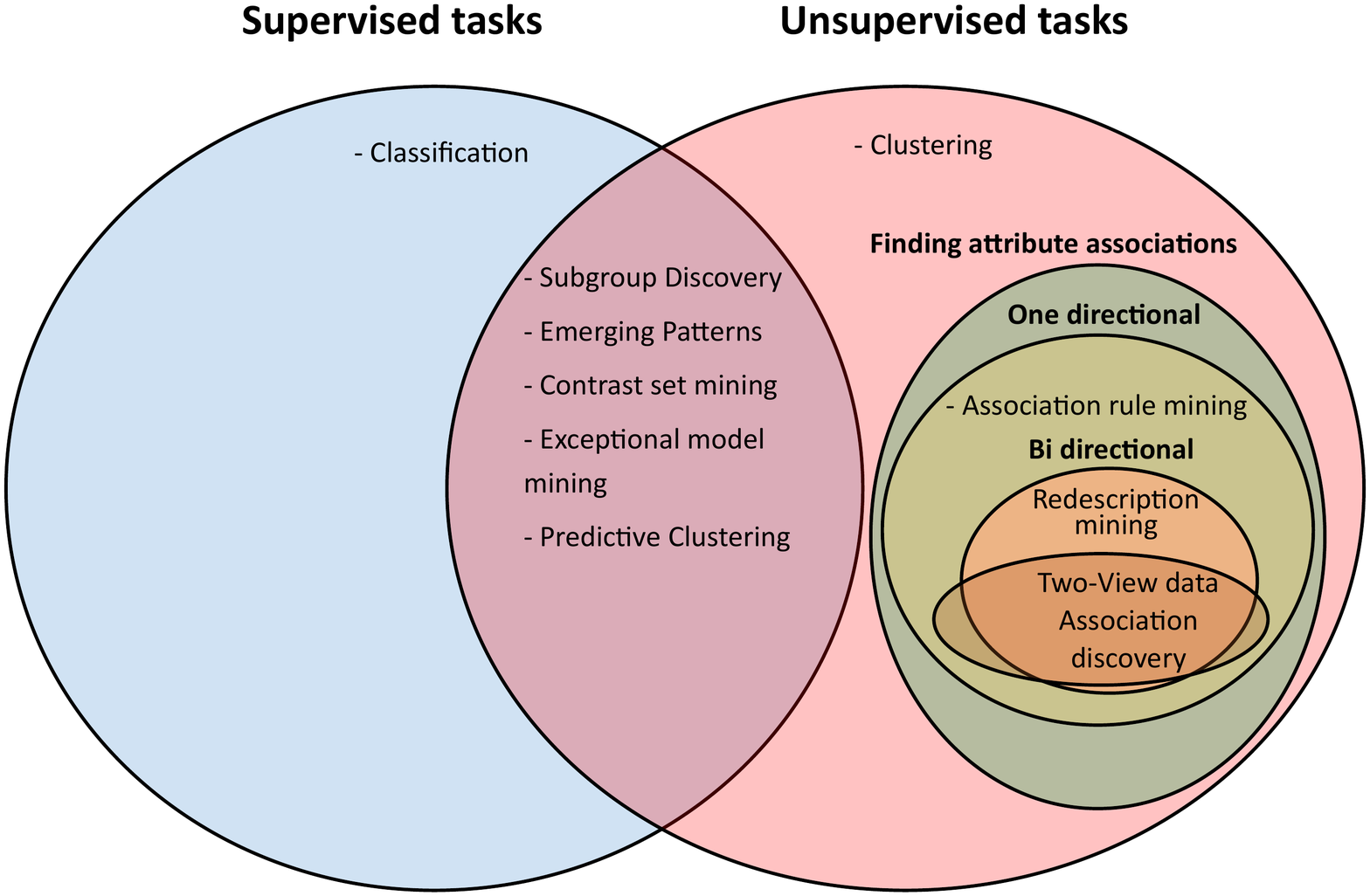}
    \caption{Relation between redescription mining and other related tasks.}
    \label{relFields}
\end{figure*}

Redescription mining is related to association rule mining \citep{AgrawalAR,HippARM,ZhangARM}, two-view data association discovery \citep{LeeuwenTWDD}, clustering \citep{CoxC,FisherC,WardC,JainClust,XuC} and it's special form conceptual clustering \citep{michalski80,FisherCC}, subgroup discovery \citep{KlosgenSD,wrobel1997SD,NovakRLSurv,HerreraSD}, emerging patterns \citep{DongEPM,NovakRLSurv}, contrast set mining \citep{BayCS,NovakRLSurv} and exceptional model mining \citep{LemanEMM}. Most important relations can be seen in Figure \ref{relFields}.

Association rule mining \citep{AgrawalAR} is related to redescription mining in the aim to find queries describing similar sets of instances which reveal associations between attributes used in these queries. The main difference is that association rules produce one directional associations while redescription mining produces bi directional associations. Two-view data association discovery \citep{LeeuwenTWDD} aims at finding a small, non - redundant set of associations that provide insight in how two views are related. Produced associations are both uni and bi directional as opposed to redescription mining that only produces bi directional connections providing interesting descriptions of instances. 

The main goal of clustering is to find groups of similar instances with respect to a set of attributes.  However, it does not provide understandable and concise descriptions of these groups which are often complex and hard to find. This is resolved in conceptual clustering \cite{michalski80,FisherCC} that finds clusters and concepts that describe them. Redescription mining shares this aim but requires each discovered cluster to be described by at least two concepts.
Clustering is extended by multi-view \citep{BickelMWC,WangMWC} and multi-layer clustering \citep{Gamberger} to find groups of instances that are strongly connected across multiple views. 

Subgroup discovery \citep{KlosgenSD,wrobel1997SD} differs from redescription mining in its goals. It finds queries describing groups of instances having unusual and interesting statistical properties on their target variable which are often unavailable in purely descriptive tasks. Exceptional model mining \citep{LemanEMM} extends subgroup discovery to more complex target concepts searching for subgroups such that a model trained on this subgroup is exceptional based on some property.  

Emerging Patterns \citep{DongEPM} aim at finding itemsets that are statistically dependent on a specific target class while Contrast Set Mining \citep{BayCS} identifies monotone conjunctive queries that best discriminate between instances containing one target class from all other instances.

\subsection{Related work in redescription mining} 
\par The field of redescription mining was introduced by \cite{RamakrishnanCart}, who present an algorithm to mine redescriptions based on decision trees, called CARTwheels. The algorithm works by building two decision trees (one for each view) that are joined in the leaves. Redescriptions are found by examining the paths from the root node of the first tree to the root node of the second. The algorithm uses multi class classification to guide the search between the two views. Other approaches to mine redescriptions include the one proposed by \cite{ZakiReas}, which uses a lattice of closed descriptor sets to find redescriptions; the algorithm for mining exact and approximate redescriptions by \cite{Parida} that uses relaxation lattice, and the greedy and the MID algorithm based on frequent itemset mining by \cite{Gallo}.
All these approaches work only on Boolean data. \par \cite{GalbrunBW} extend the greedy approach by \cite{Gallo} to work on numerical data.
Redescription mining was extended by \cite{GalbrunRel} to a relational  and by \cite{GalbrunInter} to an interactive setting. Recently, two tree-based algorithms have been proposed by \cite{Zinchenko}, which explore the use of decision trees in a non-Boolean setting and present different methods of layer-by-layer tree construction, which make informed splits at each level of the tree. \cite{Mihelcic15,Mihelcic15LNAI} proposed a redescription mining algorithm based on multi-target predictive clustering trees (PCTs) \citep{Blockeel,Kocev}. This algorithm typically creates a large number of redescriptions by executing PCTs iteratively: it uses  rules created for one view of attributes in one iteration, as target attributes for generating rules for the other view of attributes in the next iteration. A redescription set of a given size is improved over the iterations by introducing more suitable redescriptions which replace the ones that are inferior according to predefined quality criteria.

\par In this work, we introduce a redescription mining framework that allows creating multiple redescription sets of user defined size, based on user defined importance levels of  one or more redescription quality criteria. The underlying redescription mining algorithm uses multi-target predictive clustering trees \citep{Kocev} and allows the main steps of rule creation and redescription construction explained in \citep{Mihelcic15LNAI}. This is in contrast to current state of the art approaches that return all constructed redescriptions that satisfy accuracy and support constraints \citep{RamakrishnanCart,ZakiReas,Parida}, a smaller number of accurate and significant redescriptions that satisfy support constraints \citep{GalbrunBW, Zinchenko, Gallo} or optimize one redescription set of user defined size \citep{Mihelcic15LNAI}. This algorithm supports a broader process which involves the creation and effective utilization of a possibly large redescription set. 

From the expert systems perspective, the framework allows creating large and heterogeneous knowledge basis for use by the domain experts. It also allows fully automated construction of specific subsets of obtained knowledge based on predefined user-criteria. The system is modular and allows using the redescription set construction procedure as an independent querying system on the database created by merging multiple redescription sets produced by many different redescription mining approaches. Obtained knowledge can be used, for example, as a basis or complement in decision support systems. 

The framework provides means to explore and compare multiple redescription sets, without the need to expensively experiment with tuning the parameters of the underlying redescription mining algorithm. This is achieved with (i) an efficient redescription mining algorithm with a new conjunctive refinement procedure, that produces large, heterogeneous and accurate redescription sets  and (ii) redescription set construction procedure that produces one or more reduced redescription sets tailored to specific user preferences in a multi-objective optimization manner.

\par After introducing the necessary notation in Section \ref{not}, we present the framework for redescription set construction in Section \ref{framework}. First, we shortly describe the CLUS-RM algorithm, then we introduce the conjunctive refinement procedure and explain the generalized redescription set construction process. Next, we introduce the variability index: which supports a refined treatment of redescription accuracy in presence of missing values. We describe the datasets and an application involving redescription sets produced by the framework in Section \ref{applications} and perform theoretical and empirical evaluation of the framework's performance in Section \ref{evaluation}. Empirical evaluation includes quality analysis of representative sets and comparison to the set containing all discovered redescriptions, evaluation of the conjunctive refinement procedure, and quality comparison of redescriptions produced by our framework to those produced by several state of the art redescription mining algorithms, on three datasets with different properties. We conclude the paper in Section \ref{concl}. 

\section{Notation and definitions}
\label{not}
\noindent The input dataset $D=(V_1,V_2,E,W_1,W_2)$ is a quintuple of the two attribute (variable) sets ($V_1,\ V_2$), an element (instance) set $E$, and the two views corresponding to these attribute sets. Views ($W_1$ and $W_2$)  are $|E|\times |V_d|$ data matrices such that $W_{d_{i,j}}=c_k$ if an element $e_i$ has a value $c_k$ for attribute $v_j\in V_d$.

A query $q$ is a logical formula $F$ that can contain the conjunction, disjunction and negation logical operators. These operators describe logical relations between different attributes, from attribute sets $V_1$ and $V_2$,  that constitute a query. The set of all valid queries $Q$ is called a query language. The set of elements described by a query $q$, denoted $supp(q)$, is called its support. A redescription $R=(q_1, q_2)$ is defined as a pair of queries, where $q_1$ and $q_2$ contain variables from $V_1$ and $V_2$ respectively. The support of a redescription is the set of elements supported by both queries that constitute this redescription  $supp(R)=supp(q_1)\cap supp(q_2)$. We use $attr(R)$ to denote the multi-set of all occurrences of attributes in the queries of a redescription $R$. The corresponding set of attributes is denoted $attrs(R)$. The set containing all produced redescriptions is denoted $\mathcal{R}$. User-defined constraints $\mathcal{C}$ are typically  limits on various redescription quality measures.

Given a dataset $D$, a query language $Q$ over a set of attributes $V$, and a set of constraints $\mathcal{C}$, the task of redescription mining \citep{GalbrunPhd} is to find all redescriptions satisfying constraints in $\mathcal{C}$.

\subsection{Individual redescription quality measures}
\label{irqm}
\noindent The accuracy of a redescription $R=(q_1,q_2)$ is measured with the Jaccard similarity coefficient (Jaccard index).
$$J(R)=\frac{|supp(q_1) \cap supp(q_2))|}{|supp(q_1)\cup supp(q_2)|}$$

\noindent The problem with this measure is that redescriptions describing large subsets of instances often have a large intersection which results in high value of Jaccard index. As a result, the obtained knowledge is quite general and often not very useful to the domain expert. It is thus preferred to have redescriptions that reveal more specific knowledge about the studied problem and are harder to obtain by random sampling from the underlying data distribution. 

This is why we compute the statistical significance ($p$-value) of each obtained redescription. We denote the marginal probability of a query $q_1$ and $q_2$  with $p_1=\frac{|supp(q_1)|}{|E|}$ and $p_2=\frac{|supp(q_2)|}{|E|}$, respectively and the set of elements described by both as  $o=supp(q_1)\cap supp(q_2)$.
The corresponding $p$-value \citep{GalbrunPhd} is defined as 
$$pV(R)=\sum_{n=|o|}^{|E|} {|E|\choose n}(p_1\cdot p_2)^n\cdot(1-p_1\cdot p_2)^{|E|-n}$$
The $p$-value represents a probability that a subset of elements of observed size or larger is obtained by joining two random queries with marginal probabilities equal to the fractions of covered elements. It is an optimistic criterion, since the assumption that all elements can be sampled with equal probability need not hold for all datasets.

Since it is important to provide understandable and short descriptions, it is interesting to measure the number of attributes occurring in redescription queries $attr(R)$.

Below, we provide an example of a redescription, together with its associated quality measures obtained on the Bio dataset \citep{bio1,bio,GalbrunPhd}:\par \noindent
Redescription $R_{ex}'=(q_1',q_2')$ with its queries defined as: \par \noindent 
$q_1':\ (-1.8 \leq \tilde{t}_{7} \leq 4.4 \wedge 12.1 \leq \tilde{p}_{6} \leq 21.2)\ \vee\ $\par \noindent
$ (-1.6 \leq \tilde{t}_{6} \leq 1.5 \wedge 21.6 \leq \tilde{p}_{6}\leq 30.1)  $\par \noindent
$q_2':\ \text{Polarbear}$\par \noindent
describes $34$ locations which are inhabited by the polar bear. The $q_1'$ query describes the average temperature ($\tilde{t}$) and the average precipitation ($\tilde{p}$) conditions of these locations in June and July. The redescription has a Jaccard index value of $0.895$ and a $p$-value smaller than $2\cdot 10^{-16}$. The multi-set $attr(R_{ex}')=\{\tilde{t}_6,\tilde{t}_7, \tilde{p}_6, \tilde{p}_6, \text{Polarbear}\}$ and its corresponding set $attrs(R_{ex}')=\{\tilde{t}_6,\tilde{t}_7,\tilde{p}_6,\text{Polarbear}\}$. The query size of $R_{ex}'$, denoted $|attr(R_{ex}')|$, equals $5$.

\subsection{Redescription quality measures based on redescription set properties}
\label{rspqm}

\noindent We use two redescription quality measures based on properties of redescriptions contained in a corresponding redescription set.

The measure providing information about the redundancy of elements contained in the redescription support is called the average redescription element Jaccard index and is defined as: 
$$AEJ(R_i)=\frac{1}{|\mathcal{R}|-1}\cdot \sum_{j=1}^{|\mathcal{R}|} J(supp(R_i), supp(R_j)),\ i\neq j$$
 
\noindent Analogously, the measure providing information about the redundancy of attributes contained in redescription queries, called the average redescription attribute Jaccard index, is defined as: 
$$AAJ(R_i)=\frac{1}{|\mathcal{R}|-1}\cdot \sum_{j=1}^{|\mathcal{R}|} J(attrs(R_i), attrs(R_j)),\ i\neq j$$

\noindent We illustrate the average attribute Jaccard index on the redescription example from the previous subsection.
If we assume that our redescription set contains only two redescriptions $\mathcal{R}=\{R_{ex},R_{ex}'\}$ where $R_{ex}$ equals: 

\noindent $q_1:\ (-1.8 \leq \tilde{t}_{7} \leq 4.4 \wedge 12.1 \leq \tilde{p}_{6} \leq 21.2)$\par \noindent
$q_2:\ \text{Polarbear}$\par \noindent
The corresponding average attribute Jaccard index of the redescription $R_{ex}$ equals $\frac{3}{4}=0.75$ showing a high level of redundancy in the used attributes between redescription $R_{ex}$ and the only other redescription available in the set $R_{ex}'$. On the other hand, in the redescription set $\mathcal{R}=\{R_{ex}',R_{ex}''\}$, where $R_{ex}''$ contains queries:  

\noindent $q_1'':\ (7.2 \leq t_{9}^+ \leq 17.2 \wedge 13.5 \leq t_{7}^+ \leq 22.7)$\par \noindent
$q_2'':\ \text{MountainHare}$\par \noindent

\noindent the average attribute Jaccard index of the redescription $R_{ex}'$ equals $\frac{0}{7}=0$ showing no redundancy in the used attributes.

\section{Redescription mining framework}
\label{framework}
\noindent In this section, we present a redescription mining framework. It first creates a large set of redescriptions and then uses it to create one or more smaller sets that are presented to the user. This is done by taking into account the relative user preferences regarding importance of different redescription quality criteria. 

\subsection{The CLUS-RM algorihtm}
\label{clrmalg}
\noindent The framework generates redescriptions with the CLUS-RM algorithm \cite{Mihelcic15LNAI}, presented in Algorithm \ref{alg:CLUSRM}. It uses multi-target Predictive Clustering Trees (PCT) \citep{Kocev} to construct conjunctive queries which are used as building blocks of redescriptions. Queries containing disjunctions and negations are obtained by combining and transforming queries containing only conjunction operator.

\begin{algorithm}[H]
\caption{The CLUS-RM algorithm}\label{alg:CLUSRM}
\begin{algorithmic}[1]
\Require{First view data ($W_1$), Second view data ($W_2$), Constraints $\mathcal{C}$}
\Ensure{A set of redescriptions $\mathcal{R}$}
\Procedure{CLUS-RM}{}
\State  $[P_{W1init}, P_{W2init}]\leftarrow$ createInitialPCTs($W_1$, $W_2$)
\State $[r_{W1},r_{W2}]\leftarrow$ extrRulesFromPCT($P_{W1init}, P_{W2init}$)
\While{RunInd<maxIter}
\State $[D_{W1},D_{W2}]\leftarrow$ constructTargets($r_{W1}$,$r_{W2}$)
\State $[P_{W1},P_{W2}]\leftarrow$ createPCTs($D_{W1},D_{W2}$)
\State extractRulesFromPCT($P_{W1},P_{W2},r_{W1},r_{W2}$)
\State $\mathcal{R} \leftarrow \mathcal{R} \cup  \text{createRedescriptons}(r_{W_1}, r_{W_2}, \ \mathcal{C})$
\EndWhile
\State \textbf{return} $\mathcal{R}$
\EndProcedure
\end{algorithmic}
\end{algorithm}  

The algorithm is able to produce a large number of highly accurate redescriptions from which many contain only conjunction operator in the queries. This is in part the consequence of using PCTs in multi-target setting, which is known to outperform single class classification or regression trees due to the property of inductive transfer \citep{PiccartMTL}.
This distinguishes the CLUS-RM redescription mining algorithm from other state of the art solutions that in general create a smaller number of redescriptions with majority of redescription queries containing the disjunction operator.

\subsubsection{Rule construction and redescription creation}
\label{ruleredconstr}
\noindent The initial task in the algorithm is to create one PCT per view of the original data, constructed for performing unsupervised tasks, to obtain different subsets of instances (referred to as initial clusters) and the corresponding queries that describe them. To create initial clusters (line 2 in Algorithm \ref{alg:CLUSRM}), the algorithm transforms an unsupervised problem to a supervised problem by constructing an artificial instance for each original instance in the dataset. These instances are obtained by shuffling attribute values among original instances thus braking any existing correlations between the attributes. Each artificial instance is assigned a target label $0.0$ while each original instance is assigned a target label $1.0$. One such dataset is created for each view considered in the redescription mining process.
A PCT is constructed on each dataset, with the goal of distinguishing between the original and the artificial instances, and transformed to a set of rules. This transformation is achieved by traversing the tree, joining all attributes used in splits into a rule and computing its support. Each node in a tree forms one query containing the conjunction and possibly negation operators (line 3 and 7 in Algorithm \ref{alg:CLUSRM}). 

After the initial queries are created, the algorithm connects different views by assigning target labels to instances based on their coverage by queries constructed from the opposing view (line 5 in Algorithm \ref{alg:CLUSRM}). To construct queries containing attributes from $W_2$, each instance is assigned a target label $1.0$ if it is described by a query containing the attributes from $W_1$, otherwise it is assigned a value $0.0$. The process is iteratively repeated a predefined number of steps (line 4 in Algorithm \ref{alg:CLUSRM}).

Redescriptions are created as a Cartesian product of a set of queries formed on $W_1$ and a set of queries formed on $W_2$ (line 8 in Algorithm \ref{alg:CLUSRM}). All redescriptions that satisfy user defined constraints ($\mathcal{C}$): the minimal Jaccard index, the maximal $p$-value, the minimal and the maximal support are added to the redescription set. The algorithm can produce redescriptions containing conjunction, negation and disjunction operators. 

The initialization, rule construction and various types of redescription creation are thoroughly described in \citep{Mihelcic15LNAI}.

\subsubsection{Conjunctive refinement}
\label{refProc}

\noindent In this subsection, we present an algorithmic improvement to the redescription mining process presented in Algorithm \ref{alg:CLUSRM}. The aim of this method is to improve the overall accuracy of redescriptions in the redescription set by combining newly created redescriptions with redescriptions already present in redescription set $\mathcal{R}$. 

Combining existing redescription queries with an attribute by using conjunction operator has been used in greedy based redescription mining algorithms \citep{Gallo,GalbrunBW} to construct redescriptions. The idea is to expand each redescription query in turn by using a selected attribute and the selected logical operator. Such procedure, if used with the conjunction operator, leads to increase of Jaccard index but also mostly reduces the support size of a redescription. \cite{ZakiReas} combine closed descriptor sets by using conjunction operator to construct a closed lattice of descriptor sets which are used to construct redescriptions. They conclude that combining descriptor set $D_1$ and $D_2$ describing element sets $G_1$ and $G_2$ respectively, such that $G_1\subseteq G_2$, can be done by constructing a descriptor set $D_1 \cup D_2$. They conclude that the newly created descriptor set, describes the same set of elements $G_1$ as the set $D_1$. This procedure works only with attributes containing Boolean values and does not use the notion of views.

Instead of extending redescription queries with attributes connected using conjunction operator (which is usually constrained by the number of expansions), the conjunctive refinement procedure compares support of each redescription $R=(q_1,q_2)$ in the redescription set with the selected redescription $R_{ref}=(q'_1,q'_2)$. It merges the queries of these two redescriptions with the $\{\wedge\}$ operator to obtain a new redescription $R_{new}=(q_1\wedge q_1',q_2\wedge q_2')$ if and only if $supp(R)\subseteq supp(R_{ref})$.
We extend and prove the property described in \cite{ZakiReas} in a more general setting, combining redescriptions with arbitrary type of attributes and a finite amount of different views. We demonstrate how to use it efficiently with numerical attributes and show that this procedure does not decrease the accuracy of a redescription. In fact, if $\exists e\in E,\ e\in supp(q_1) \ \vee\ \exists e'\in E,\ e'\in supp(q_2)$ such that $e\notin supp(q'_1) \vee\ e'\notin supp(q'_2)$, than $J(R_{new})>J(R)$.

If the attributes contain numerical values, we can transform the redescription $R_{ref}$, given an arbitrary redescription $R\in \mathcal{R}$ such that $supp(R)\subseteq supp(R_{ref})$, to redescription $R_{ref}'=(q''_1,q''_2)$ such that $R_{ref}'$ has tighter numerical bounds on all attributes contained in the queries, $supp(R)\subseteq supp(R_{ref}')$ and that $J(supp(R),supp(R_{ref}'))\geq J(supp(R),supp(R_{ref}))$. By doing this, we increase the probability of finding the element $e$ or $e'$ as described above, which leads to improving the accuracy of redescription $R_{new}$. The construction procedure of such redescription is explained in Section S1.1 (Online Resource 1).
The redescription $R_{ref}'$ is used as a refinement redescription when numerical attributes are present in the data.

\par \vspace{2mm}\noindent We can now state and prove the following lemma: 

\begin{lemma}
For every redescription $R\in \mathcal{R}$, for every redescription $R_{ref}=(q_1',q_2')$, where $q_1'= q_{a_1}\ \wedge\ q_{a_2}\ \wedge\ \dots\ \wedge\ \ q_{a_n}$, $a_i\in attrs(R_{ref}),\ \forall i\in\{1,\dots,n\}$ and $n\in \mathbb{N}$, $q_2'= q_{b_1}\ \wedge\ q_{b_2}\ \wedge\ \dots\ \ \wedge\  q_{b_m}$, $b_j\in attrs(R_{ref}),\ \forall j \in \{1,\dots,m\}$ and $m\in \mathbb{N}$. If $supp(R)\subseteq supp(R_{ref})$ then for a redescription $R_{new}=(q_1\wedge q_1', q_2\wedge q_2')$ it holds that $J(R_{new})\geq J(R)$ and $supp(R_{new})=supp(R)$.
\end{lemma}

\noindent The proof of Lemma 3.1 for redescription mining problems containing two views can be seen in Section S1.1 (Online Resource 1).  General formulation with $n$ arbitrary views is proven by mathematical induction. It is easily seen from the proof that if $\exists e\in E,\ e\in supp(q_1) \ \vee\ \exists e'\in E,\ e'\in supp(q_2)$ such that $e\notin supp(q'_1) \vee\ e'\notin supp(q'_2)$ then $supp(q_1\wedge q_1') \cup supp(q_2\wedge q_2')\subset supp(q_1) \cup supp(q_2)$ thus ultimately $J(R_ {new})>J(R)$.

\par
\begin{figure}[ht]
    \centering
    \includegraphics[width=0.5\textwidth]{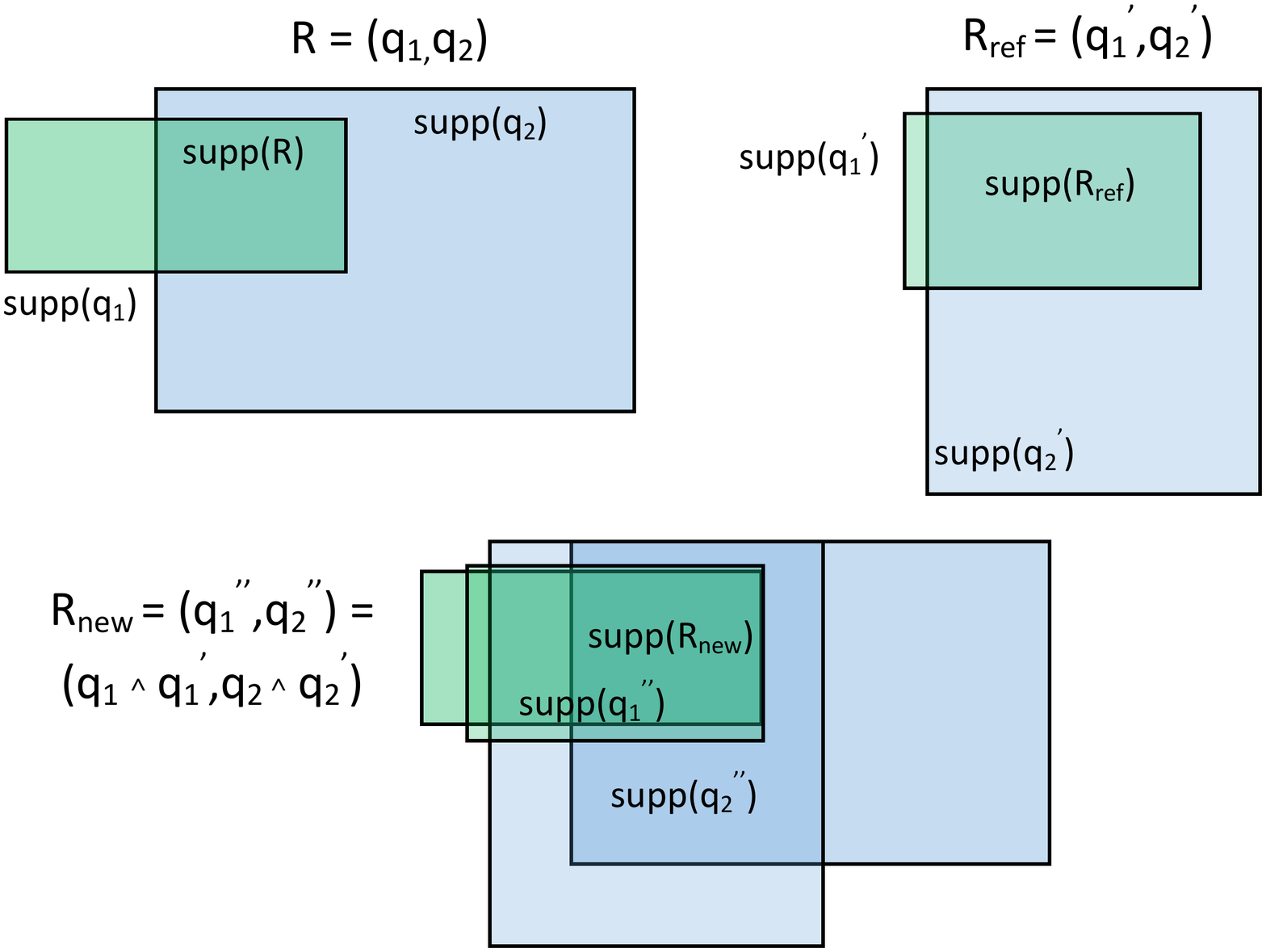}
    \caption{Demonstration of the effects of the conjunctive refinement on a support of the improved redescription and corresponding redescription queries. For the supports represented on the figure it holds:  $supp(R)\subset supp(R_{ref})$. As a consequence:  $supp(R)=supp(R_{new})$, $J(R_{new})> J(R)$.}
    \label{refinement}
\end{figure}

 \noindent The conjunctive refinement is demonstrated in Figure \ref{refinement}.

\par \vspace{2mm} \noindent Line 8 from Algorithm \ref{alg:CLUSRM} is replaced with the procedure \emph{$\mathcal{R}\leftarrow$createAndRefineRedescriptions($rw_1,rw_2,\mathcal{R}, \mathcal{C}$)} which is presented in Algorithm \ref{alg:ref}.

\begin{algorithm}[ht]
\caption{The redescription set refinement procedure}\label{alg:ref}
\begin{algorithmic}[1]
\Require{Rules created on $W_1$ ($rw_1$), Rules created on $W_2$ ($rw_2$), Redescription set $\mathcal{R}$, Constraints $\mathcal{C}$}
\Ensure{A set of redescriptions $\mathcal{R}$}
\Procedure{ConstructAndRefine}{}
\For{$R_{new}\in rw_1\times rw_2$}
\If{$R_{new}.J\geq\mathcal{C}.minJref$}
\For{$R\in \mathcal{R}$}
\State $R.Refine(R_{new})$
\State $R_{new}.Refine(R)$
\EndFor
\If{$R_{new}.J\geq\mathcal{C}.minJ$}
\State $\mathcal{R}\leftarrow \mathcal{R} \cup R_{new}$
\EndIf
\EndIf
\EndFor
\State \textbf{return} $\mathcal{R}$
\EndProcedure
\end{algorithmic}
\end{algorithm}  

\noindent The procedure described in Algorithm \ref{alg:ref} and demonstrated in Figure S1 applies conjunctive refinement by using redescriptions that satisfy the user defined constraints $\mathcal{C}$ and redescriptions that satisfy looser constraints on the Jaccard index ($R.J\geq C.minRefJ$, $C.minRefJ\leq C.minJ$). These constraints determine the amount and variability of redescriptions used to improve the redescription set.

The refinement procedure, in combination with redescription query minimization explained in \cite{Mihelcic15LNAI}, provides grounds for mining more accurate yet compact redescriptions.

\subsection{Generalized redescription set construction}
\label{GSC}

\begin{figure}[ht]
    \centering
    \includegraphics[width=0.5\textwidth]{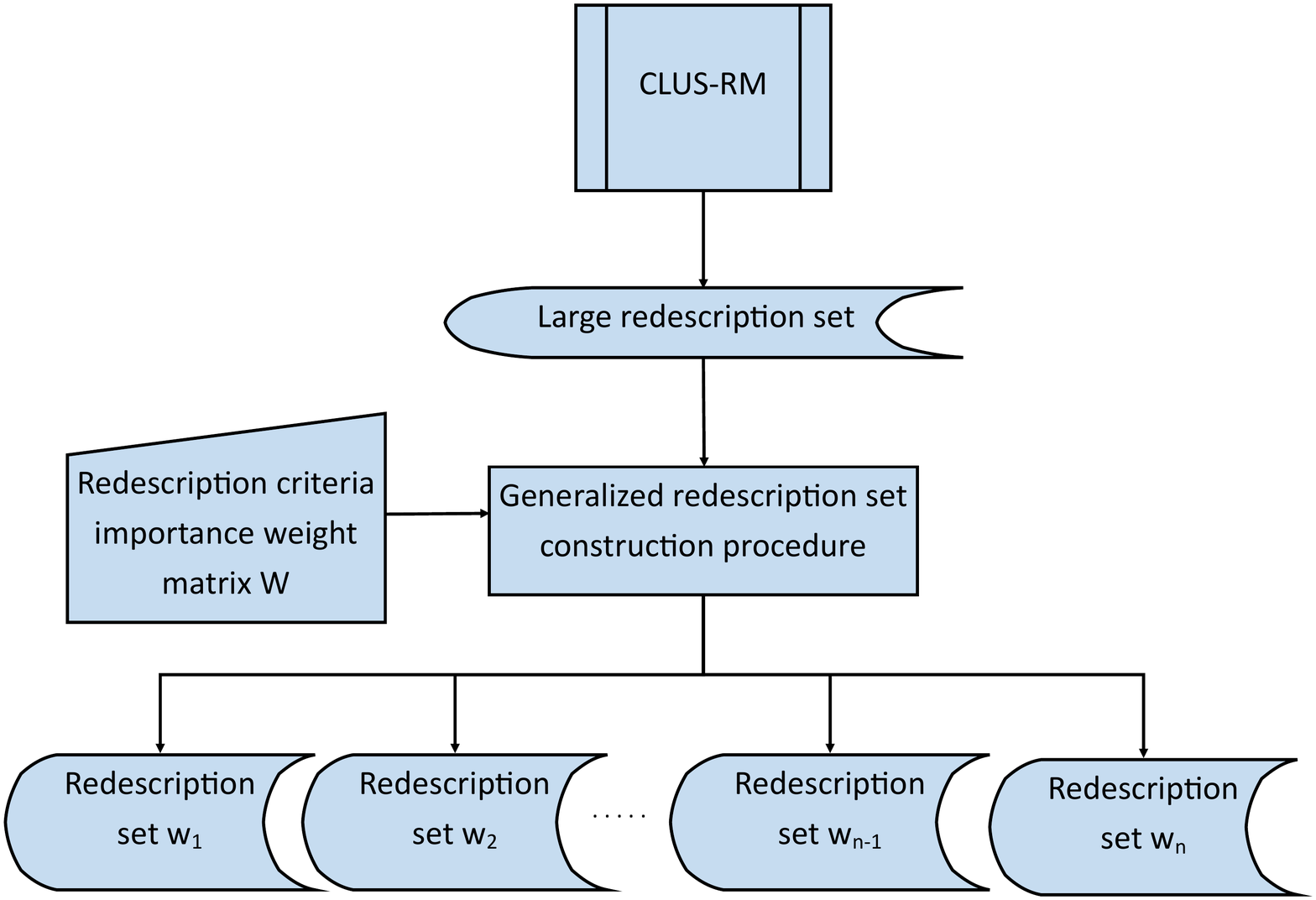}
    \caption{Flowchart representing the redescription set construction process.}
    \label{fig:grscProcFlow}
\end{figure}

\noindent The redescription set obtained by Algorithm \ref{alg:CLUSRM} contains redescriptions satisfying hard constraints described in the previous subsections. It is often very large and hard to explore. For this reason, we extract one or more smaller sets of redescriptions that satisfy additional preferential properties on objective redescription evaluation measures, set up by the user, and present them for exploration. This process is demonstrated in Figure \ref{fig:grscProcFlow}. 

Producing summaries and compressed rule set representations is important in many fields of knowledge discovery. In the field of frequent itemset mining such dense representations include closed itemsets \citep{PasquierCISAR} and free sets \citep{BoulicautFC}. The approaches using set pattern mining construct a set by enforcing constraints on different pattern properties, such as support, overlap or coverage \citep{GunsPSM}. Methods developed in information theory consider sets that provide the best compression of a larger set of patterns. These techniques use properties like the Information Bottleneck \citep{TishbyIBM} or the Minimum description length \citep{GrunwaldMDLP}. The work on statistical selection of association rules developed by \cite{BoukerSRS} presented techniques to eliminate irrelevant rules based on dominance, which is computed on several possibly conflicting criteria. If some rule is not strictly dominated by any other rule already in the set, the minimal similarity with some representative rule is used to determine if it should be added to the set.  

Redescriptions are highly overlapping with respect to described instances and attributes used in the queries. It is often very hard to find fully dominated redescriptions, and the number of dominated redescriptions that can be safely discarded is relatively small compared to a set of all created redescriptions. Our approach, to create a set of user defined (small) size, does not use a representative rule to compute the similarity. Instead, it adds redescriptions to the final redescription set by using the scalarization technique \citep{caramia2008multi} developed in multi-objective optimization to find the optimal solution when faced with many conflicting criteria. If the corresponding optimization function is minimized, given positive weights, the solution is a strict pareto optimum, otherwise it is a weak pareto optimum \citep{caramia2008multi} of a multi objective optimization problem. Similar aggregation technique is used in multi attribute utility theory - MAUT \citep{Winterfeldt1975} to rank the alternatives in decision making problems. 

Each redescription is evaluated with a set of criteria known from the literature or defined by the user. The final quality score is obtained by aggregating these criteria with user-defined importance weights to produce a final numerical score. Based on this score, the method selects one non-dominated redescription, based on utilised quality criteria, at each step of redescription set construction.

The procedure generalizes the current redescription set construction approaches in two ways: 
1) it allows defining importance weights to different redescription quality criteria and adding new ones to enable constructing redescription sets with different properties which provides different insight into the data, 2) it allows creating multiple redescription sets by using different weight vectors, support levels, Jaccard index thresholds or redescription set sizes. Thus, it in many cases eliminates the need to make multiple runs of a redescription mining algorithm. 

One extremely useful property of the procedure is that it can be used by any existing redescription mining algorithm, or a combination thereof. In general, larger number of diverse, high quality redescriptions allows higher quality reduced sets construction.

Are there any elements in the data that share many common properties? Can we find a subset of elements that allows multiple different redescriptions? Can we find very diverse but accurate redescriptions? What is the effect of reducing redescription query size to the overall accuracy on the observed data? What are the effects of missing values to the redescription accuracy? What is our confidence that these redescriptions will remain accurate if missing values are added to our set? This is only a subset of questions that can be addressed by observing redescription sets produced by the proposed procedure. The goal is not to make redescription mining subjective in the sense of interestingness \citep{TuzhilinInterest} or unexpectedness \citep{PadmanabhanUnexp}, but to enable exploration of mined patterns in a more versatile manner.

The input to the procedure is a set of redescriptions produced by Algorithm \ref{alg:CLUSRM} and an importance weight matrix defined by the user. The rows of the importance weight matrix define the users' importance for various redescription quality criteria. The procedure creates one output redescription set for each row in the importance weight matrix (line 3 in Algorithm \ref{alg:extRed}).
The procedure works in two parts: first it computes element and attribute occurrence in redescriptions from the original redescription set (line 2 in Algorithm \ref{alg:extRed}). This information is used to find the redescription that satisfies the user defined criteria and describes elements by using attributes that are found in a small number of redescriptions from the redescription set. When found (line 4 in Algorithm \ref{alg:extRed}), it is placed in the redescription set being constructed (line 5 in Algorithm \ref{alg:extRed}). Next, the procedure iteratively adds non-dominated redescriptions (lines 7-9 in Algorithm \ref{alg:extRed}) until the maximum allowed number of redescriptions is placed in the newly constructed set (line 6 in Algorithm \ref{alg:extRed}).

\begin{algorithm}[H]
\caption{Generalized redescription set construction}\label{alg:extRed}
\begin{algorithmic}[1]
\Require{Redescription set $\mathcal{R}$, Importance weight matrix $\mathcal{W}$, Size of reduced set $n$}
\Ensure{A set of reduced redescription sets $\mathcal{R}_{red}$}
\Procedure{ReduceSet}{}
\State  $[E_{ocur}, A_{ocur}]\leftarrow$ computeCoocurence($\mathcal{R}$)
\For{$w_{i}\in \mathcal{W}$}
\State $R_{first}\leftarrow$ findSpecificRed($\mathcal{R},E_{cooc},A_{cooc}, w_{i}$)
\State $\mathcal{R}_{w_i}\leftarrow \mathcal{R}_{w_i} \cup R_{first}$
\While{$|R_{w_i}|<n$}
\State $R_{best}\leftarrow$ findBest($\mathcal{R},\mathcal{R}_{w_i},w_{i}$)
\State $\mathcal{R}_{w_i}\leftarrow \mathcal{R}_{w_i} \cup R_{best}$
\EndWhile
\State $\mathcal{R}_{red}\leftarrow \mathcal{R}_{red}\cup \{\mathcal{R}_{w_i}\}$
\EndFor
\State \textbf{return} $\mathcal{R}_{red}$
\EndProcedure
\end{algorithmic}
\end{algorithm}  

In the current implementation, we use $6$ redescription quality criteria, however more can be added. Five of these criteria are general redescription quality criteria, the last one is used when the underlying data contains missing values and will be described in the following section.

The procedure \emph{findSpecificRed} uses the information about the redescription Jaccard index, $p$-value, query size and the occurrence of elements described by the redescription and attributes found in redescriptions queries in redescriptions from the redescription set. 
The $p$-value quality score of a redescription $R$ is computed as: 
\[   score_{pval}(R)=\left\{
\begin{array}{ll}
      \frac{log_{10}(pV(R))}{17}+1 &, pV(R) \geq 10^{-17}\\
      0 &,  pV(R) < 10^{-17} \\
\end{array} 
\right. \]
\noindent The logarithm is applied to linearise the $p$-values and the normalization $17$ is used because $10^{-17}$ is the smallest possible $p$-value that we can compute. 

The element occurrence score of a redescription is computed as: $score_{ocurEl}(R)=\frac{\sum_{e_k\in supp(R)} E_{ocur}[k]}{\sum_{j=1}^{|E|} E_{ocur}[j]}$. The attribute occurrence score is computed in the same way as: $score_{ocurAt}(R)=\frac{\sum_{a_k\in attrs(R)} A_{ocur}[k]}{\sum_{j=1}^{|V_1|+|V_2|} A_{ocur}[j]}$. We also compute the score measuring query size in redescriptions: 

\[   score_{size}=\left\{
\begin{array}{ll}
      \frac{|attr(R)|}{k} &, |attr(R)|< k\\
      1 &, k\leq |attr(R)| \\
\end{array} 
\right. \]

\noindent The user-defined constant $k$ denotes redescription complexity normalization factor. In this work we use $k=20$, because redescriptions containing more than $20$ variables in the queries are highly complex and hard to understand.

The first redescription is chosen by computing:
$R_{first}=arg min_{R}\ ( w_0\cdot (1.0-J(R))+w_1\cdot score_{pval}(R) +w_2\cdot score_{ocurEl}(R) +w_3 \cdot score_{ocurAt}(R) + w_4 \cdot score_{size}(R) )$. Each following redescription is evaluated with a score function that computes redescription similarity to each redescription contained in the redescription set. The similarity is based on described elements and attributes used in redescription queries. This score thus allows controlling the level of redundancy in the redescription set. For a redescription $R_i\in \mathcal{R}\backslash \mathcal{R}_{red}$ we compute: $score_{elemSim}(R_i)= max_{j}\ J(supp(R_i),supp(R_j)) ,\ j=1,\dots, |\mathcal{R}_{red}|$ and $score_{attrSim}(R)=max_{j}\ J(attrs(R_i),attrs(R_j)) ,\ j=1,\dots, |\mathcal{R}_{red}|$. 

Several different approaches to reducing redundancy among redescriptions have been used before, however no exact measure was used to select redescriptions or to assess the overall level of redundancy in the redescription set. \cite{ZakiReas} developed an approach for non-redundant redescription generation based on a lattice of closed descriptor sets, \cite{RamakrishnanCart} used the parameter defining the number of times one class or descriptor is allowed to participate in a redescription. This is used to make a trade-off between exploration and redundancy. \cite{Parida} computed non-redundant representations of sets of redescriptions containing some selected descriptor (set of Boolean attributes). \cite{GalbrunBW} defined a minimal contribution parameter each literal must satisfy to be incorporated in a redescription query. This enforces control over redundancy on the redescription level. Redundancy between different redescriptions is tackled in the Siren tool \cite{GalbrunSiren} as a post processing (filtering) step. \cite{Mihelcic15LNAI} use weighting of attributes occurring in redescription queries and element occurrence in redescription supports based on work in subgroup discovery \citep{GambergerL02SDW,Lavrac:2004SDMLR}. 

We combine the redescription $p$-value score with its support to first add highly accurate, significant redescriptions with smaller support, and then incrementally add accurate redescriptions with larger support size. 
Candidate redescriptions are found by computing:  $R_{best}=arg min_{R}\ ( w_0\cdot (1.0-J(R))+w_1\cdot (\frac{k}{n}\cdot score_{pval}(R)+(1-\frac{k}{n})\cdot \frac{supp(R)}{|E|}) +w_2\cdot score_{elemSim}(R) +w_3 \cdot score_{attrSim}(R) + w_4 \cdot score_{size}(R) )$, where $k$ denotes the number of redescriptions contained in the set under construction at this step.

\subsection{Missing values}
\label{missing}

\noindent There are more possible ways of computing the redescription Jaccard index when the data contains missing values. The approach that assumes that all elements from redescription support containing missing values are distributed in a way to increase the redescription Jaccard index is called optimistic ($J_{opt}$). Similarly, the approach that assumes that all elements from redescription support containing missing values are distributed in a way to decrease the redescription Jaccard index is called pessimistic ($J_{pess}$). The rejective Jaccard index evaluates redescriptions only by observing elements that do not contain missing values for attributes contained in redescription queries.  These measures are discussed in \citep{GalbrunBW}. 
The Query non-missing Jaccard index ($J_{qnm}$), introduced in \citep{Mihelcic15LNAI}, is an approach that gives a more conservative estimate than the optimistic Jaccard index but more optimistic estimate than the pessimistic Jaccard index. The main evaluation criteria for this index is that a query (containing only the conjunction operator) can not describe an element that contains missing values for attributes in that query. This index is by its value closer to the optimistic than the pessimistic Jaccard index. However, as opposed to the optimistic approach, redescriptions evaluated by this index contain in their support only elements that have defined values for all attributes in redescription queries and that satisfy query constraints. The index does not penalize the elements containing missing values for attributes in both queries which are penalized in the pessimistic Jaccard index. 

In this paper, we introduce a natural extension to the presented measures: the redescription variability index. This index measures the maximum possible variability in redescription accuracy due to missing values. This allows finding redescriptions that have only slight variation in accuracy regardless the actual value of the missing values. It also allows reducing very strict constraints imposed by the pessimistic Jaccard index that might lead to the elimination of some useful redescriptions.

\par \vspace{2mm}\noindent The redescription variability index is defined as: 
$variability(R) = J_{opt}(R)-J_{pes}(R)$.

\noindent Formal definitions of pessimistic and optimistic Jaccard index can be seen in Section S1.2 (Online resource 1).

\par \vspace{2mm}\noindent The scores used to find the first and the best redescription in generalized redescription set construction (Section \ref{GSC}) are extended to include the \emph{variability} score.

\noindent Our framework optimizes query non-missing Jaccard but reports all Jaccard index measures when mining redescriptions on the data containing missing values. In principle with the generalized redescription set construction, we can return reduced sets containing accurate redescriptions found with respect to each Jaccard index. Also, with the use of variability index, the framework allows finding redescriptions with accuracy affected to a very small degree by the missing values which is not possible by other redescription mining algorithms in the literature. The only approach working with missing values ReReMi requires preforming multiple runs of the algorithm to make any comparisons between redescriptions mined by using different version of Jaccard index.

\section{Data description and applications}
\label{applications}

\noindent We describe three datasets used to evaluate CRM-GRS and demonstrate its application on a Country dataset.

\subsection{Data description}

\label{datDesc}

\noindent The evaluation and comparisons are performed on three datasets with different characteristics: the Country dataset \citep{unctad,wb,Gamberger}, the Bio dataset \citep{bio1,bio,GalbrunPhd} and the DBLP dataset \citep{dblp,GalbrunPhd}. Detailed description of each dataset can be seen in Section S2 (Online resource 1).

\begin{table}[H]
\centering
\caption{Description of datasets used to perform experiments}
    \begin{tabular}{ | p{2cm} | p{25mm} | p{25mm} |}
    \hline
    \textbf{Dataset} & \textbf{$W_1$ attributes} & \textbf{$W_2$ attributes} \\ \hline
    \textbf{Country} \par \noindent $|E| = 199$ \par \noindent countries & Numerical ($49$) \par \noindent World Bank \par \noindent Year: 2012 \par \noindent Country info & Numerical ($312$) \par \noindent UNCTAD \par \noindent Year: 2012 \par \noindent Trade Info  \\ \hline
    \textbf{Bio} \par \noindent $|E|=2575$ \par \noindent geographical locations & Numerical ($48$) \par \noindent Climate conditions  & Boolean $(194)$\par \noindent mammal species  \\ \hline
    \textbf{DBLP} \par \noindent $|E|=6455$\par \noindent authors & Boolean ($304$) \par \noindent author-conference bi-partite graph & Boolean ($6455$)\par \noindent co-authorship network\\
    \hline
    \end{tabular}
    \label{tab:data}
\end{table}

\noindent Descriptions of all attributes used in the datasets are provided in the document (Online Resource 2). 

\subsection{Application on the Country dataset}

The aim of this study is to discover regularities and interesting descriptions of world countries with respect to their trading properties and general country information (such as various demographic, banking and health related descriptors). We will focus on redescriptions describing four European countries: Germany, Czech Republic, Austria and Italy, discovered as a relevant cluster in a study performed by \cite{Gamberger}. This study investigated country and trade properties of EU countries with potential implications to a free trade agreement with China. This or similar use-case may be a potential topic of investigation for economic experts but the results of such analysis could also be of interest to the policymakers and people involved in export or import business.

First step in the exploration process involves specifying various constraints on produced redescriptions. Determining parameters such as minimal Jaccard index or minimal support usually requires extensive experimentation. These experiments can be performed with CRM-GRS with only one run of redescription mining algorithm by using minimal Jaccard index of $0.1$, minimal support of $5$ countries (if smaller subsets are not desired) and $p$-value of $0.01$. Parameters specifying reduced set construction can now be tuned to explore different redescription set sizes, minimal Jaccard thresholds or minimal and maximal support intervals. Results of such meta analysis (presented in Section S2.2.2 (Online resource 1)) show little influence of setting minimal Jaccard threshold on this dataset, however right choice of minimal support is important. Redescription sets using minimal support threshold of $5$ countries show superior properties and may contain useful knowledge. 

\begin{figure}[ht]
    \centering
    \includegraphics[width=0.5\textwidth]{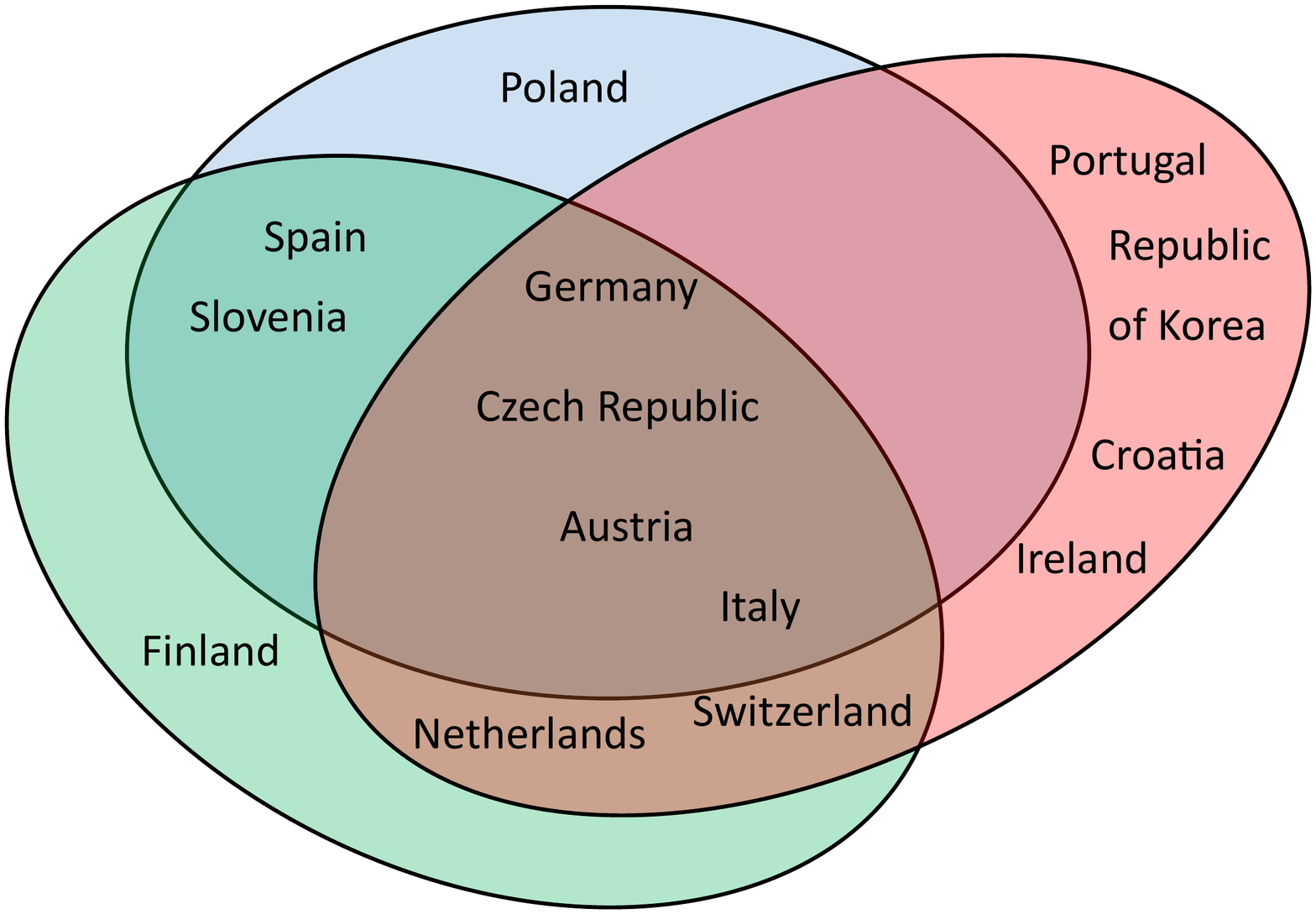}
    \caption{Similarities between different, mostly European, countries.}
    \label{fig:ApplicationsEx}
\end{figure}

We present three different redescriptions describing specified countries and revealing their similarity to several other countries (demonstrated in Figure \ref{fig:ApplicationsEx}).

\noindent Redescriptions $R_{blue},\ R_{green}$ and $R_{red}$ are defined as: 

\par \noindent $qb_1:\ 13.2 \leq \text{ POP}_{14} \leq 15.2\ \wedge\ 3.1 \leq MORT \leq 5.0 $ \par \noindent
$ \ \wedge\   0.0 \leq \text{ POP\_GROWTH } \leq 0.5$ \par \noindent
$qb_2:\  13.2 \leq \text{E/I\_MiScManArt} \leq 15.2 \ \wedge\ 28.0 \leq \text{E\_MedSTehInMan} \leq 40.0$. 
\par \noindent ($J_{qnm}(R_{blue})=J_{opt}(R_{blue})=1.0$, $J_{pess}(R_{blue})=0.88$, $pV(R_{blue})=2.3\cdot 10^{-10}$, $|supp(R_{blue})|=7$)

\par \noindent $qg_1:\ 16.2 \leq \text{ POP}_{64} \leq 21.1\ \wedge\ 2.9 \leq MORT \leq 4.5 $ \par \noindent
$ \ \wedge\   16.2 \leq \text{ RUR\_POP } \leq 50.1 \ \wedge\  0.2 \leq \text{W\_REM} \leq 1.4$ \par \noindent
$qg_2:\ 0.8 \leq \text{E/I\_ElMachApp} \leq 1.8 \ \wedge\ 93.0 \leq \text{E\_AlocProd} \leq 99.0\ \wedge\ 1.1\leq \text{E/I\_SpecMach}\leq 4.3$.  
\par \noindent ($J_{qnm}(R_{green})=J_{opt}(R_{green})=J_{pess}(R_{green})=1.0$, $pV(R_{green})=1.9\cdot 10^{-11}$, $|supp(R_{blue})|=9$)

\par \noindent $qr_1:\ 3.6 \leq \text{ MORT } \leq 4.7\ \wedge\ 22.9 \leq \text{CRED\_COV} \leq 100.0 $ \par \noindent
$ \ \wedge\   77.3 \leq \text{ M2} \leq 238.9$ \par \noindent
$qr_2:\ 0.1 \leq \text{E/I\_Cereals} \leq 1.7 \ \wedge\ 1.2 \leq \text{E/I\_BevTob} \leq 3.1\ \wedge\ 0.7 \leq \text{E/I\_SpecMach}\leq 4.3 $. 
\par \noindent ($J_{qnm}(R_{red})=J_{opt}(R_{red})=1.0$, $J_{pess}(R_{red})=0.45$, $pV(R_{red})=6.3\cdot 10^{-12}$, $|supp(R_{red})|=10$)

\begin{table}[h]
\centering
\caption{Description of attributes from $R_{blue},\ R_{green}$ and $R_{red}$}
    \begin{tabular}{ | p{28mm} | p{50mm} |}
    \hline
    \textbf{Code} & \textbf{Description} \\ \hline
     $POP_{14}$& \% of population aged [0,14]\\ \hline
     $POP_{64}$&  \% of population aged 65+\\ \hline
     $MORT$ & Mortality under $5$ years per $1000$ \\ \hline
     $POP\_GROWTH$ & \% of population growth \\ \hline
     $RUR\_POP$ & \% of population living in rural area \\ \hline
     $W\_REM$ & \% of GDP spent on worker's remittances and compensation\\ \hline
     $CRED\_COV$ & \% of adults listed by private credit bureau \\ \hline
     $M_2$ & \% of GDP as (quasi) money \\ \hline
     $E,\ I,\ E/I$ & export, import, export to import ratio \\ \hline
     $MiScManArt$ & Miscellaneous manufactured articles \\ \hline
     $MedSTehInMan$ & Medium - skill, technology - intensive manufactures \\ \hline
     $ElMachApp$ & Electrical machinery, apparatus and appliances \\ \hline
     $AlocProd$ & All allocated products \\ \hline
     $SpecMach$ & Specialised machinery \\ \hline
     $Cereals$ & Cereals and cereal preparations \\ \hline
     $BevTob$ & Beverages and tobacco \\ \hline
    \end{tabular}
    \label{tab:mapping}
\end{table}

Presented redescriptions (attribute descriptions available in Table \ref{tab:mapping}) confirm several findings reported in \citep{Gamberger}. Mainly, high export of medium - skill and technology - intensive manufactures, export of beverages and tobacco, low percentage of young population. Additionally, these redescriptions reveal high percentage of elderly population (age $65$ and above), lower (compared to world average of $47.4$) but still present mortality rate of children under $5$ years of age (per $1000$ living) and small to medium percentage of rural population. The credit coverage (percentage of adults registered for having unpaid depths, repayment history etc.) varies between countries but is no less than $20\%$ adult population. The money and quasi money (M2 - sum of currency outside banks etc.) is between substantial $77.3\%$ and very large $239\%$ of total country's GDP. For additional examples see Section S2.2.3, Figure S11 (Online resource 1).

Output of CRM-GRS can be further analysed with visualization and exploration tools such as the Siren \citep{GalbrunSiren} (available at \url{http://siren.gforge.inria.fr/main/}) or the InterSet \citep{MihelcicInterSet} (available at \url{http://zel.irb.hr/interset/}). In particular, the InterSet tool allows exploration of different groups of related redescriptions, discovery of interesting associations, multi-criteria filtering and redescription analysis on the individual level.

\section{Evaluation and comparison}
 \label{evaluation}
\noindent In this section we present the results of different evaluations. First, we perform a theoretical comparison of our approach with other state of the art solutions which, includes description of advantages and drawbacks of our method. Next, we apply the generalized redescription set construction procedure to these datasets starting from redescriptions created by the CLUS-RM algorithm. We evaluate the conjunctive refinement procedure and perform a thorough comparison of our reduced sets with the redescription sets obtained by several state of the art redescription mining algorithms. The comparisons use measures on individual redescriptions (Section \ref{irqm}) as well as  measures on redescription sets (Section \ref{rspqm}). We also use the normalized query size defined in Section \ref{GSC}.

The execution time analysis, showing significant time reduction when using generalized redescription set construction instead of multiple CLUS-RM runs, is described in Section S2.4 (Online resource 1).

\subsection{Theoretical algorithm comparison}

We compare the average case time and space complexity of the CRM-GRS with state of the art approaches and present the strengths and weaknesses of our framework.

\begin{table}[h]
\centering
\caption{Time and space complexity of redescription mining algorithms and the generalized redescription set construction procedure}
    \begin{tabular}{ | p{2cm} | p{25mm} | p{25mm} |}
    \hline
    \textbf{Algorithm} & Time comp. & Space comp. \\ \hline
    \textbf{CRM-GRS}&  $O(z\cdot (|V_1|+|V_2|)\cdot |E|^2 + z^2\cdot |E|)$ (No refinement)\par \noindent $O(z\cdot (|V_1|+|V_2|)\cdot |E|^2 + z^3\cdot |E|)$ (refinement) &  $O(z)$ \Tstrut\Bstrut  \\ \hline
    \textbf{CARTWh.} & $O(z\cdot (|V_1|+|V_2|)\cdot |E|^2)$ & $O(z)$ \Tstrut\Bstrut\\
    \hline
    \textbf{Split trees} & $O(z\cdot (|V_1|+|V_2|)\cdot |E|^2)$ & $O(z)$ \Tstrut\Bstrut \\
    \hline
    \textbf{Layered trees} &  $O(z\cdot (|V_1|+|V_2|)\cdot |E|^2)$ & $O(z)$ \Tstrut\Bstrut \\
    \hline
    \textbf{Greedy} & $O(|V_1|\cdot |V_2|\cdot |E|)$ & $O(1)$ \Tstrut\Bstrut\\
    \hline
    \textbf{MID} & $O(|\mathcal{C}|\cdot |E|\cdot 2^l)$ & $O(1)$ \Tstrut\Bstrut\\
    \hline
    \textbf{Closed Dset} & $O(|\mathcal{C}|\cdot |E|\cdot 2^l)$ & $O(|\mathcal{C}|)$ \Tstrut\Bstrut\\ \hline
    \textbf{Relaxation Latt.} & $max(O(|\mathcal{B}|\cdot log$ $(|E|)+(|V_1|+|V_2|)\cdot |E|$, $O(L\cdot log(|E|)+(|V_1|+|V_2|)\cdot |E|))$  & $O(|\mathcal{B}|)$ \Tstrut\Bstrut\\ \hline \hline
      \textbf{GRSC} & $O(|\mathcal{R}|\cdot |E|)$ & $O(|\mathcal{R}|)$   \Tstrut\Bstrut  \\ \hline
    \end{tabular}
    \label{tab:complexCmp}
\end{table}

\noindent The term $z = 2^d-1$ in Table \ref{tab:complexCmp} denotes the number of nodes in the tree and is constrained by the tree depth $d$. $\mathcal{C}$ denotes the set of produced maximal closed frequent itemsets, $l$ denotes the length of the longest itemset, $\mathcal{B}$ a set of produced biclusters, $L=\sum_{c\in \mathcal{B}} |c|$ and $\mathcal{R}$ denotes a set of produced redescriptions.

We can see from Table \ref{tab:complexCmp} that the CRM-GRS has slightly higher computational complexity than other tree - based approaches (which is based on time complexity of algorithm C4.5), caused by complexity of underlying redescription mining algorithm CLUS-RM. Optimizations proposed in \citep{Mihelcic15LNAI} lower average time complexity of basic algorithm to $O(z\cdot (|V_1|+|V_2|)\cdot |E|^2$ and algorithm with refinement to $O(z\cdot (|V_1|+|V_2|)\cdot |E|^2 + z^2\cdot |E|)$. Worst-case complexity with the use of refinement is $O(z\cdot (|V_1|+|V_2|)\cdot |E|^2 + z^4\cdot |E|)$. It is the result of a very optimistic estimate that produced redescriptions satisfying user constraints grow quadratically with the number of nodes in the tree (this is only the case if no constraints on redescriptions are enforced). In reality, it has at most linear growth. Furthermore, term $z^2\cdot |E|$ is only dominating if $z>(|V_1|+|V_2|)\cdot |E|$. Since redescription queries become very hard to understand if they contain more than $10$ attributes, even with $2$ attributes in each of two views, this term is dominated when  $|E|>255$ instances.

Greedy approaches \citep{Gallo,GalbrunBW} are less affected by the increase in number of instances than the tree-based approaches, but are more sensitive to the increase in number of attributes.

Complexity of approaches based on closed and frequent itemset mining \citep{Gallo,ZakiReas} depends on the number of produced frequent or closed itemsets which in worst case equals $2^{|V_1|+|V_2|}$. Similarly, the complexity of approach proposed by \cite{Parida} depends on the number of created biclusters and their size.

One property of our generalized redescription set construction procedure (GRSC) is that it can be used to replace multiple runs of expensive redescription mining algorithms. Analysis from Table \ref{tab:complexCmp} and in S2.6 (Online resource 1) shows that it has substantially lower time complexity than all state of the art approaches except the MID and the Closed Dset. However, even for this approaches, it might be beneficial to use GRSC instead of multiple runs of these algorithms when $|\mathcal{C}|\cdot 2^l>|\mathcal{R}|$.   

Since a trade-off between space and time complexity can be made for each of the analysed algorithms, we write the space complexity as a function of stored itemsets, rules, redescriptions or clusters. To reduce execution time, these structures can be stored in memory together with corresponding instances which increases space complexity to $O(C_{old}\cdot |E|)$ for all approaces.

One drawback of our method is increased memory consumption ($O(z^2)$ in the worst case). Since we memorize all distinct created redescriptions that satisfy user constraints, it is among more memory expensive approaches. Although, the estimate $O(z^2)$ is greatly exaggerated, and is in real applications at most $O(z)$, it is currently the only approach that memorizes and uses all created redescriptions to create diverse and accurate redescription sets for the end users.
If memory limit is reached, we use the GRCS procedure (called in line 8 of  Algorithm \ref{alg:CLUSRM}) to create reduced redescription sets of predefined properties. Only redescriptions from these sets are retained allowing further execution of the framework.

Greedy and the MID approaches are very memory efficient since they store only a small number of candidate redescriptions in memory. Other tree-based approaches store two decision trees at each iteration, Closed Dset \citep{ZakiReas} approach saves a closed lattice of descriptor sets and the relaxation lattice approach \citep{Parida} saves produced biclusters.  

The main advantages of our approach are that it produces a large number of diverse, highly accurate redescriptions which enables our multi-objective optimization procedure to generate multiple, high quality redescription sets of differing properties that are presented to the end user.

\subsection{Experimental procedure}
\label{expproc}

\noindent In this section we explain all parameter settings used to perform evaluations and comparisons with various redescription mining algorithms. 

For all algorithms, we used the maximal $p$-value threshold of $0.01$ (the strictest significance threshold). The minimal Jaccard index was set to $0.2$ for the DBLP dataset based on results presented in \cite{GalbrunPhd}, Table 6.1, p. 46. The same is set to $0.6$ for the Bio dataset based on results in \cite{GalbrunPhd} Table 7, p. 301. The threshold $0.5$ for the Country dataset was experimentally determined. Minimal support was set to $10$ elements for the DBLP, based on \cite{GalbrunPhd} p.48, and the same is used for the Bio dataset. Country dataset is significantly smaller thus we set this threshold to $5$ elements. 
Impact of changing minimal Jaccard index and minimal support is data dependant. Increasing these thresholds causes a drop in diversity of produced redescriptions, resulting in high redundancy and in some cases inadequate number of produced redescriptions. However, it also increases minimal and average redescription Jaccard index and support size. Lowering these thresholds has the opposite effect, increasing diversity but potentially reducing overall redescription accuracy or support size. Increasing maximal $p$-value threshold allows more redescriptions (although less significant) to be considered as candidates for redescription set construction. The effects of changing minimal Jaccard index and minimal support size on the produced redescription set of size $50$ by our framework on Country, Bio and DBLP dataset can be seen in Section S2.2.2 (Online resource 1).

We compared the CLUS-RM algorithm with the generalized redescription set construction procedure (CRM-GRS), to the ReReMi, the Split trees and the Layered trees algorithms implemented in the tool called Siren \citep{GalbrunSiren}. The specific parameter values used for each redescription mining algorithm can be seen in Section S2 (Online Resource 1).

\subsection{Analysis of redescription sets produced with CRM-GRS}
\label{setQ}

\noindent We analyse a set containing all redescriptions produced by CLUS-RM algorithm (referred to as a \emph{large set of redescriptions}) and the corresponding sets of substantially smaller size constructed from this set by generalized redescription set construction procedure (referred to as \emph{reduced sets of redescriptions}) on three different datasets. 

For the purpose of this analysis, we create redescriptions without using the refinement procedure and disallow multiple redescriptions describing the same set of instances. To explore the influence of using different importance weights on properties of produced redescription sets, we use the different weight combinations given in Table \ref{tab:WMat}. 

\begin{table}[ht!]
\caption{A matrix containing different combinations of importance weights for the individual redescription quality criteria.}
$$W=\begin{bmatrix}
    \text{J}  & \text{pV} & \text{AJ}& \text{EJ} & RQS & RV \\
    0.2       & 0.2 & 0.2 & 0.2 & 0.2 & 0.0 \\
    0.4       & 0.2 & 0.1 & 0.1 & 0.2 & 0.0 \\
    0.6	      & 0.2 & 0.0 & 0.0 & 0.2 & 0.0 \\
    0.0       & 0.2 & 0.3 & 0.3 & 0.2 & 0.0 
\end{bmatrix}$$
\label{tab:WMat}
\end{table}

\noindent In the rows $1,2$ and $3$ of matrix $W$, we incrementally increase the importance weight for the Jaccard index and equally decrease the weight for the element and attribute Jaccard index in order to explore the effects of finding highly accurate redescriptions at the expense of diversity. The last row explores the opposite setting that completely disregards accuracy and concentrates on diversity.

By using importance weights in each row of matrices $W$ (Table \ref{tab:WMat}) and $W_{miss}$ (Table \ref{tab:WMat1}), we create redescription sets containing $25,\ 50,\ 75,\ 100,$ $\ 125,\ 150,\ 175$ and $200$ redescriptions. We plot the change in element/attribute coverage, average redescription Jaccard index, average $p$-value, average element/attribute Jaccard index and average query size against the redescription set size. Information about redescriptions in the large set is used as a baseline and compared to the quality of reduced sets. 

\subsubsection{The analysis on the Bio dataset}

\begin{figure*}[ht!]
\captionsetup[subfloat]{farskip=1mm}
    \centering
\subfloat{%
  \includegraphics[width=0.4\textwidth]{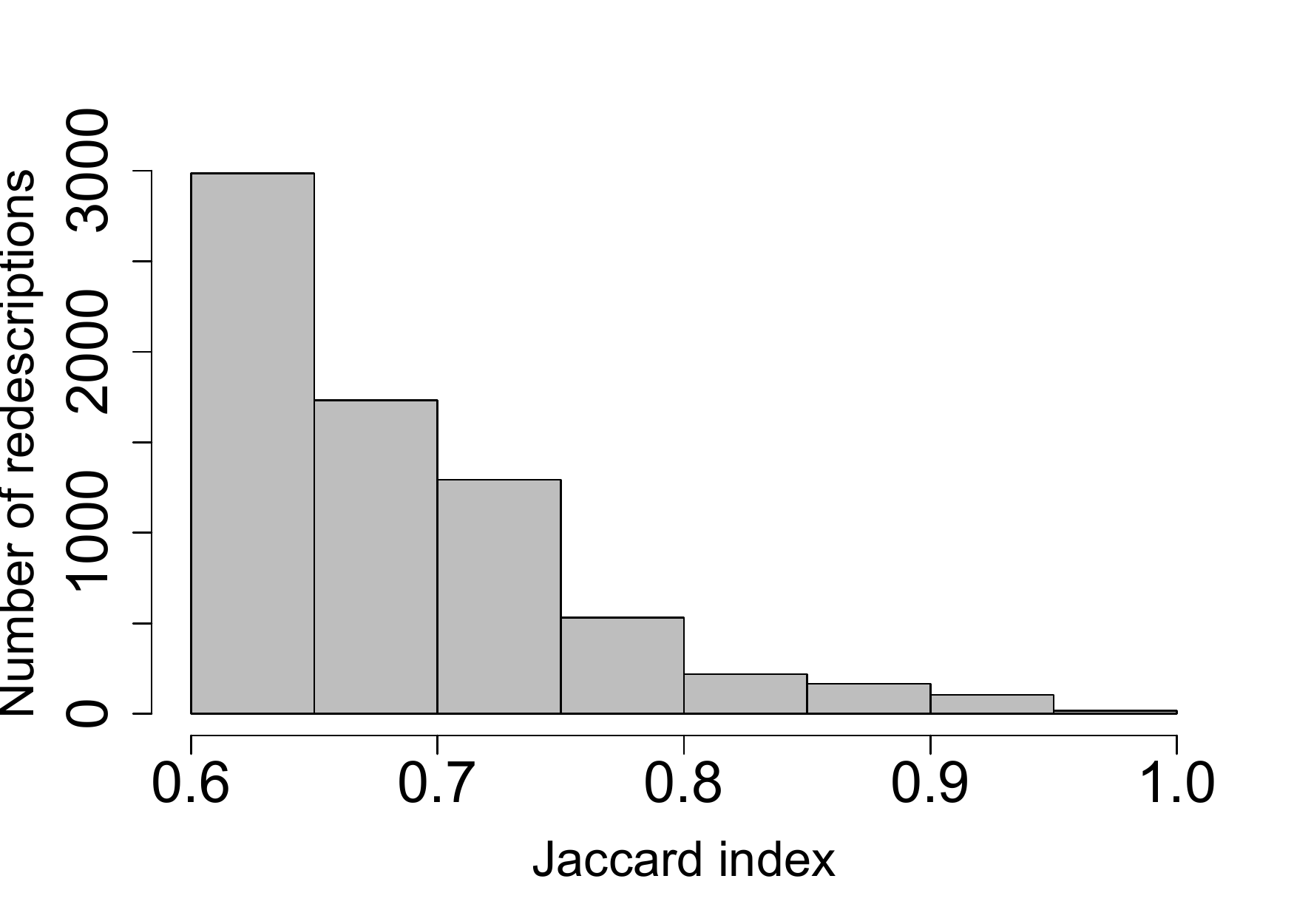}
}\qquad
\subfloat{%
    \includegraphics[width=0.4\textwidth]{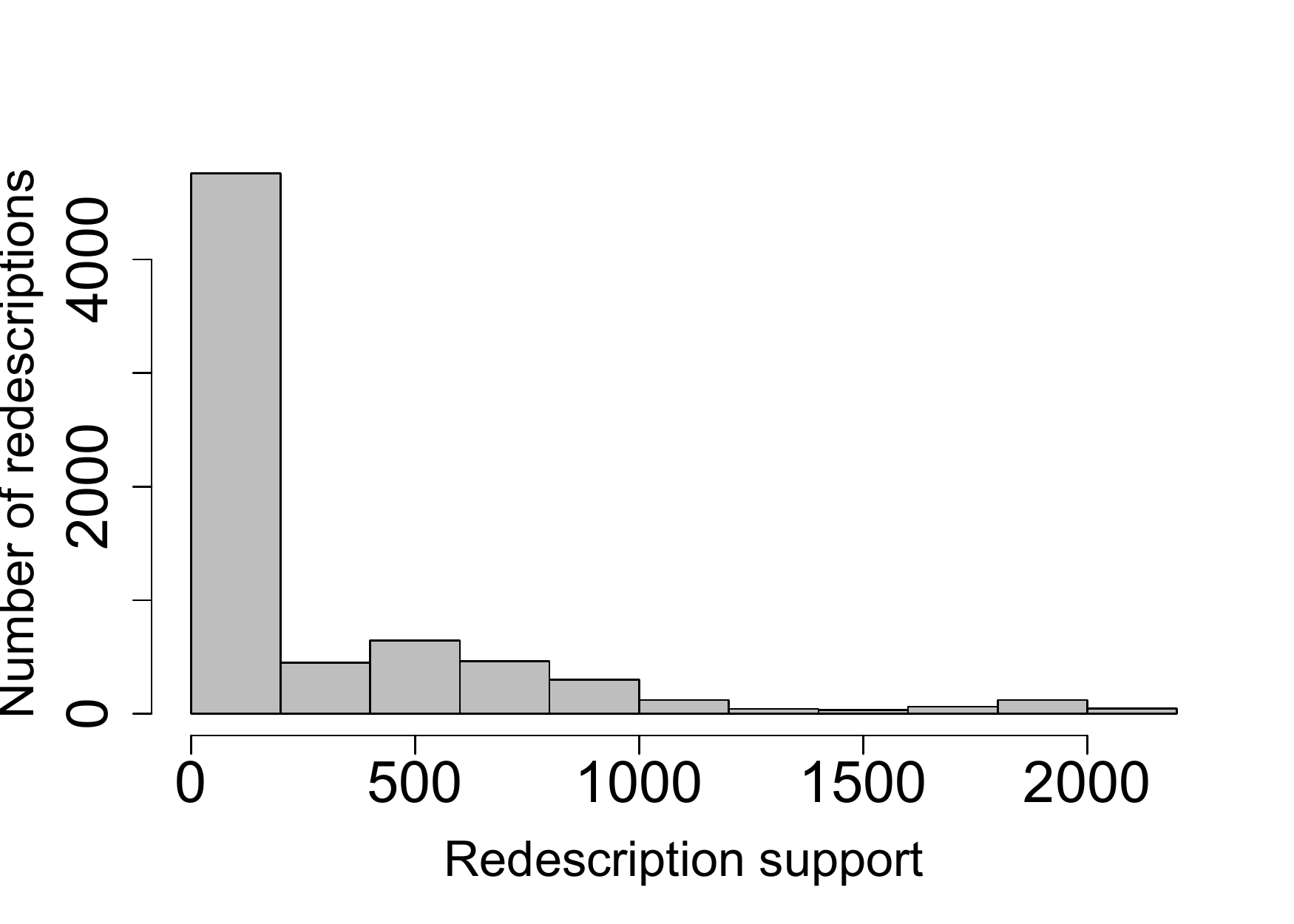}
  }\qquad
  \subfloat{%
  \includegraphics[width=0.4\textwidth]{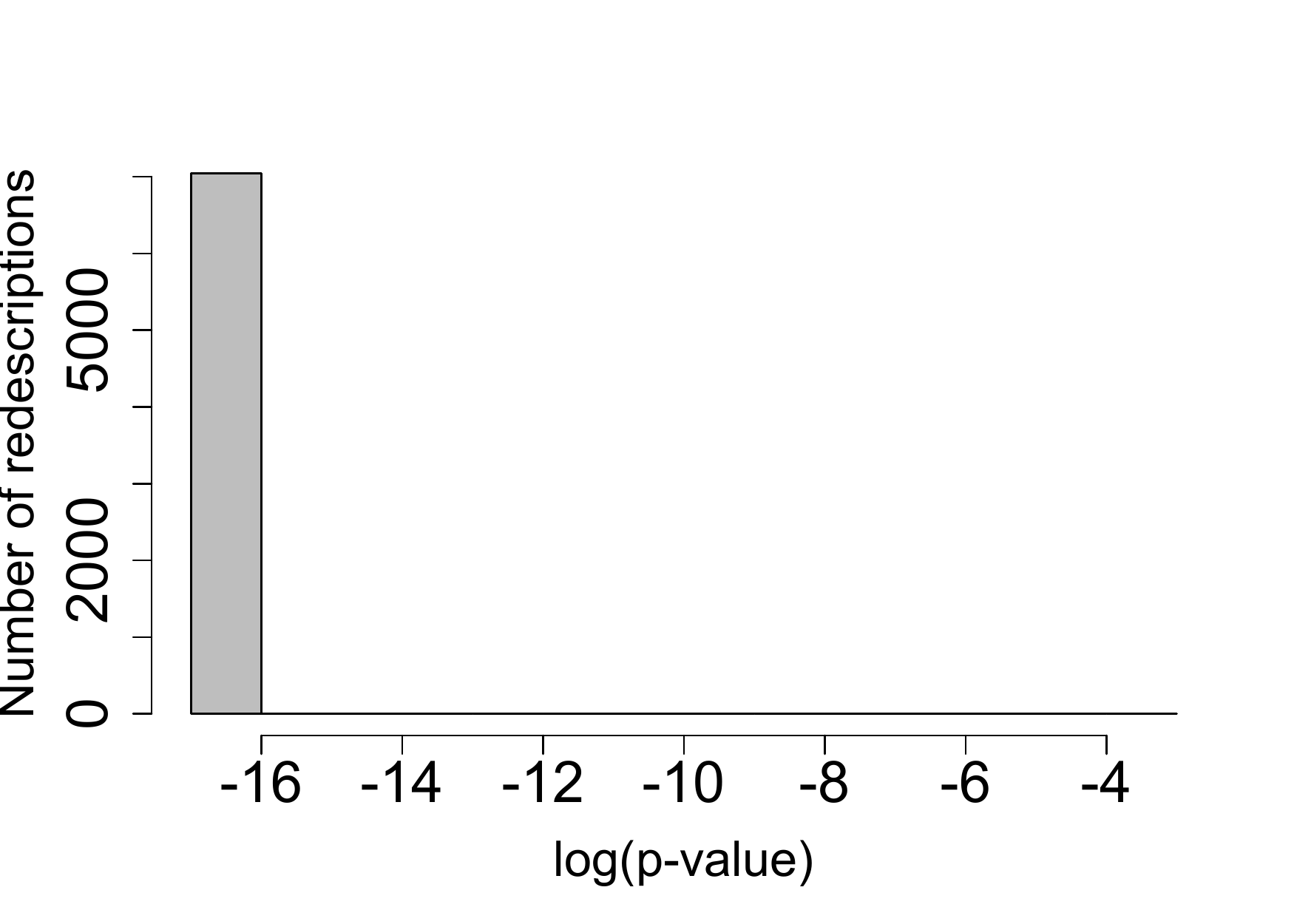}
  }\qquad
  \subfloat{%
    \includegraphics[width=0.4\textwidth]{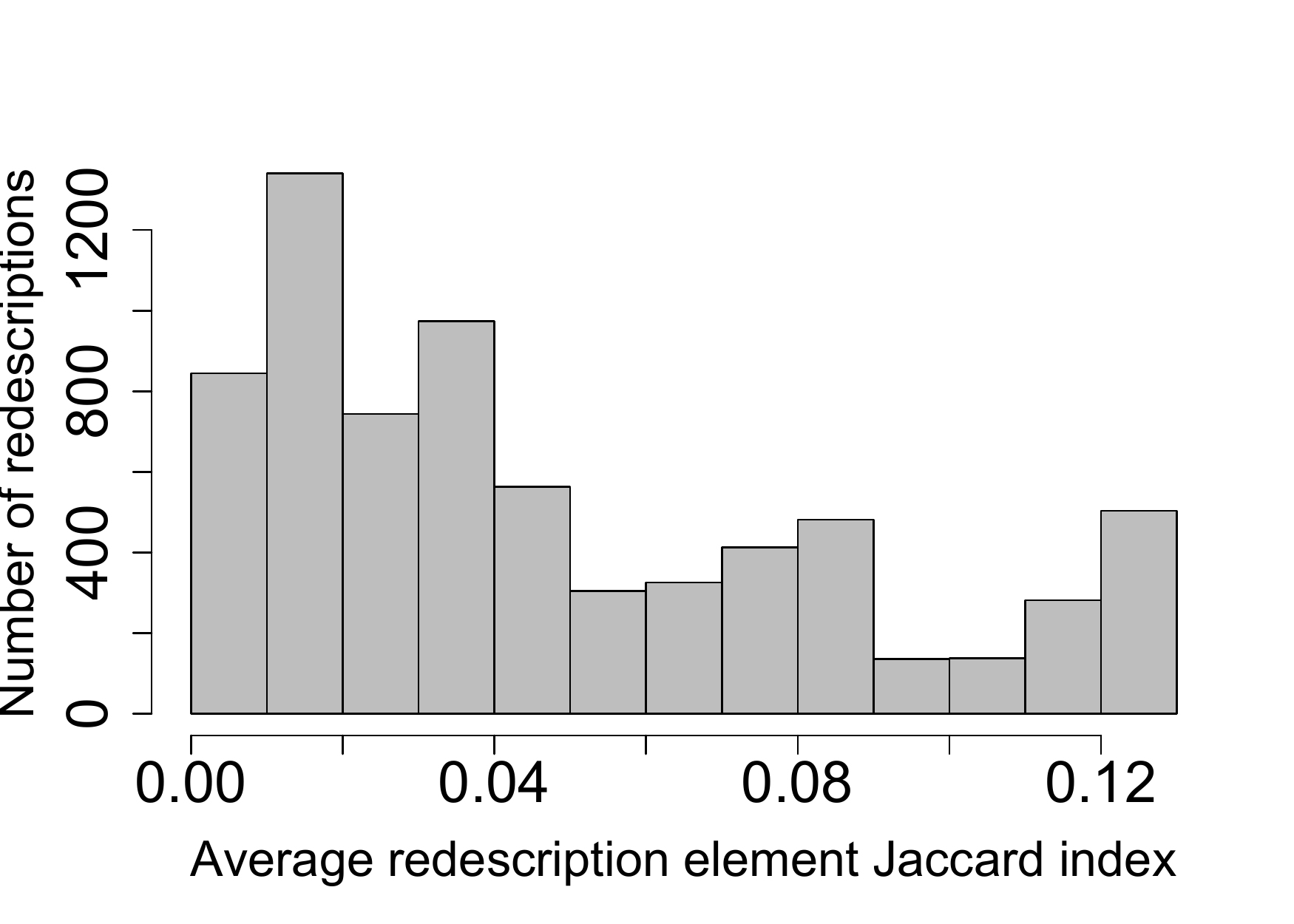}
  }\qquad
  \subfloat{%
  \includegraphics[width=0.4\textwidth]{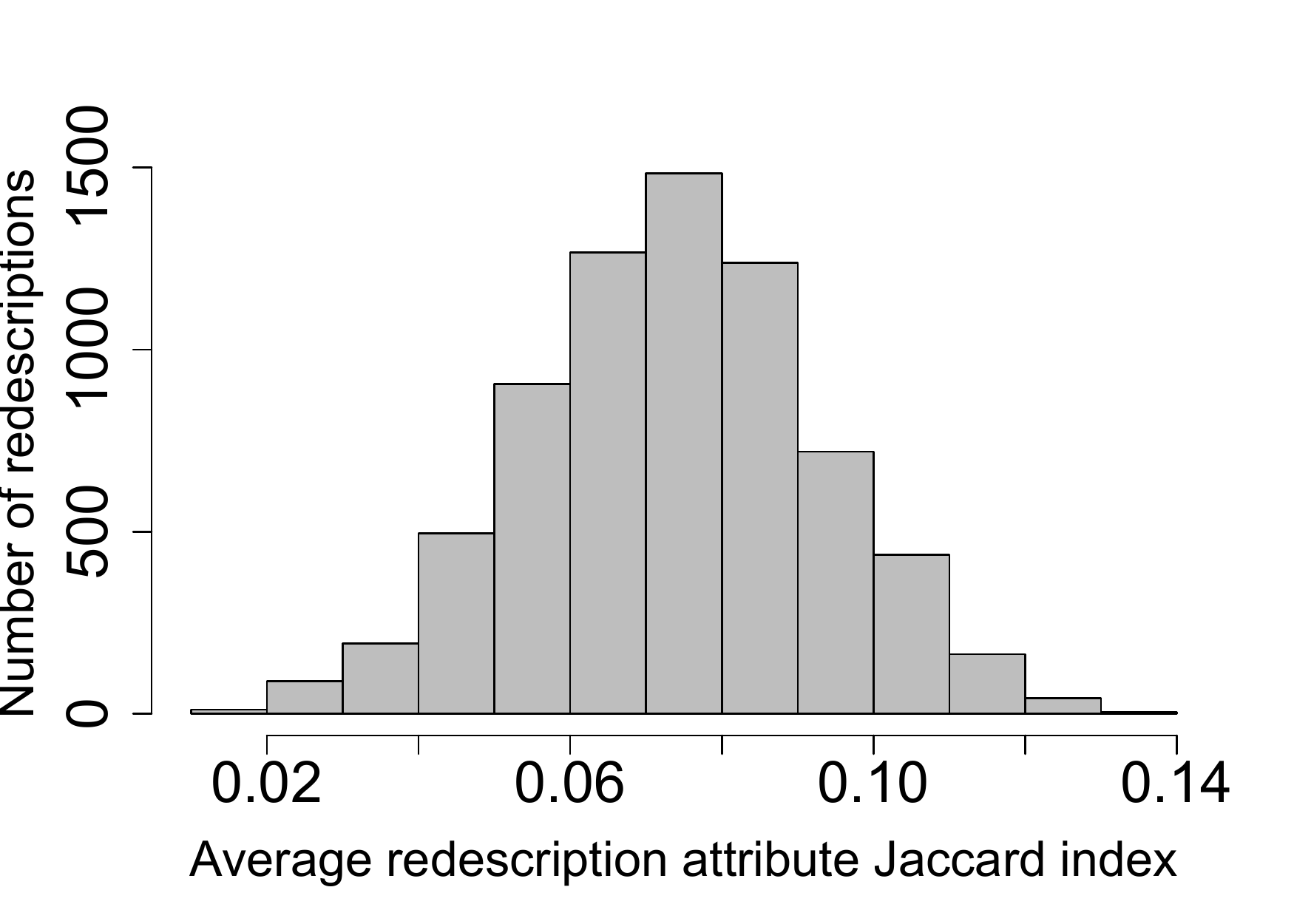}
}\qquad
\subfloat{%
    \includegraphics[width=0.4\textwidth]{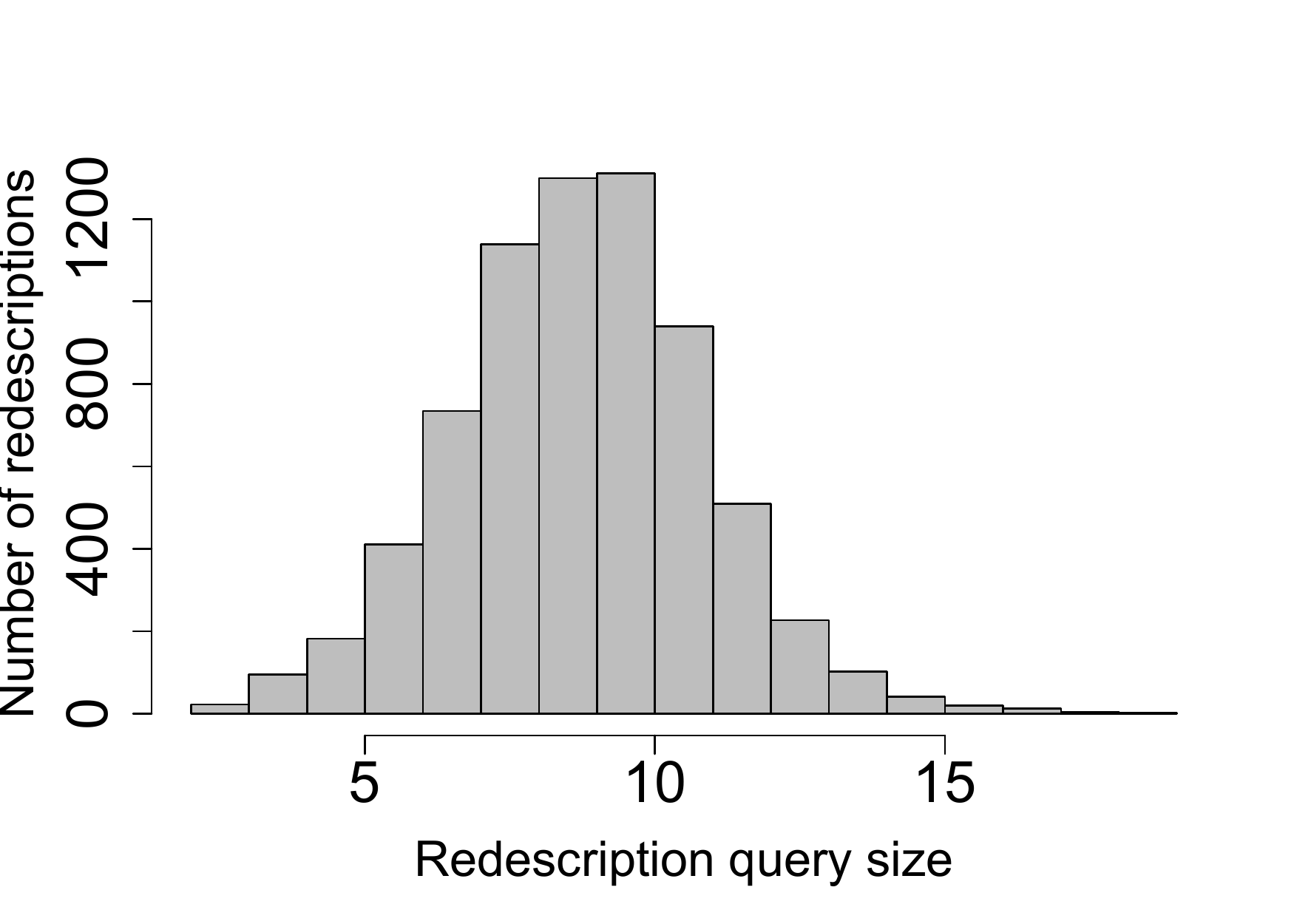}
}
  \caption{Histograms showing distributions of different redescription quality measures for the large redescription set containing $7413$ redescriptions. Redescriptions are created on the Bio dataset.}
   \label{fig:RSAnal}
\end{figure*}

\begin{figure*}[p!]
\captionsetup[subfloat]{farskip=0.01mm}
    \centering
\subfloat{%
  \includegraphics[width=0.42\textwidth]{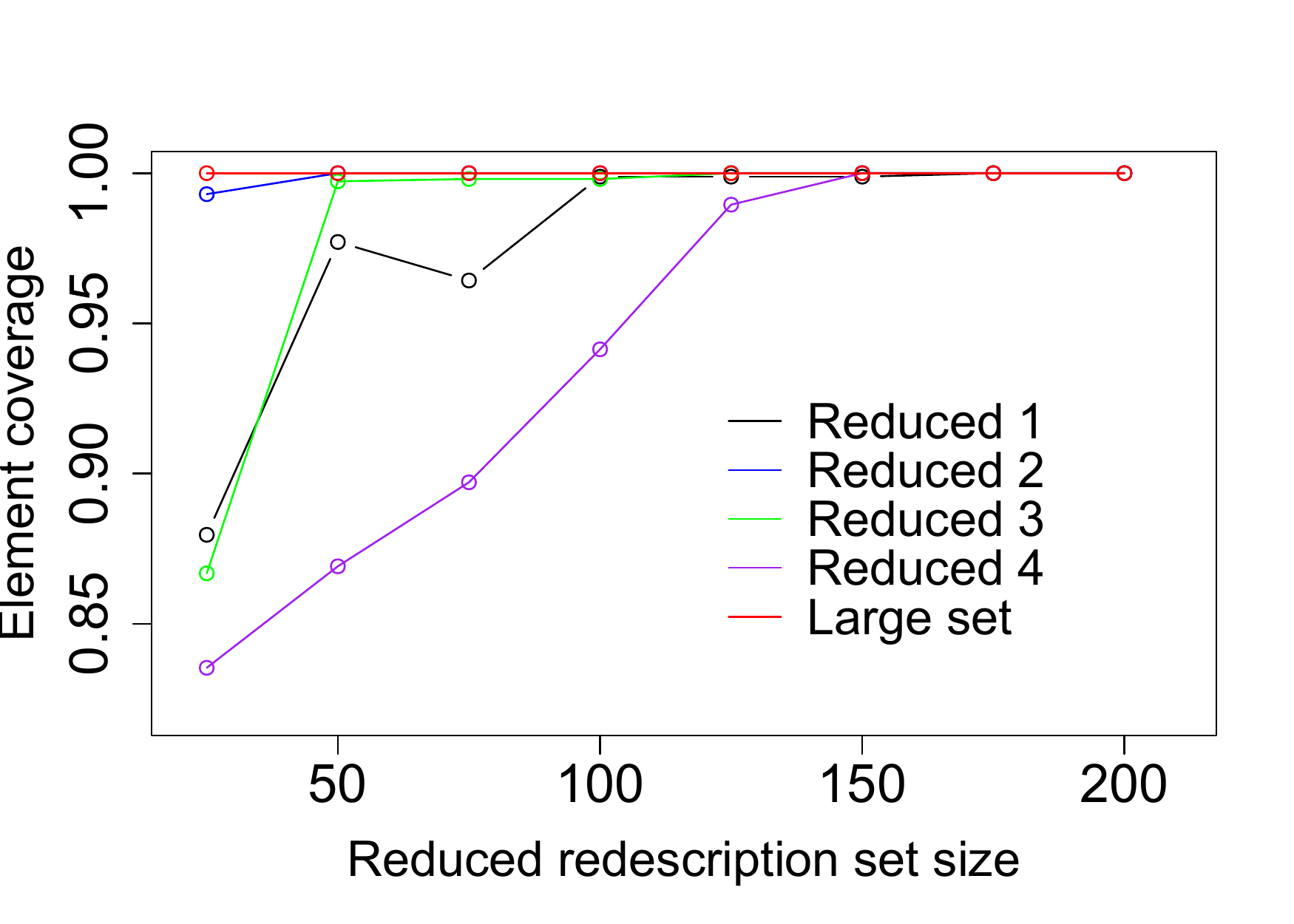}
}\qquad
\subfloat{%
    \includegraphics[width=0.42\textwidth]{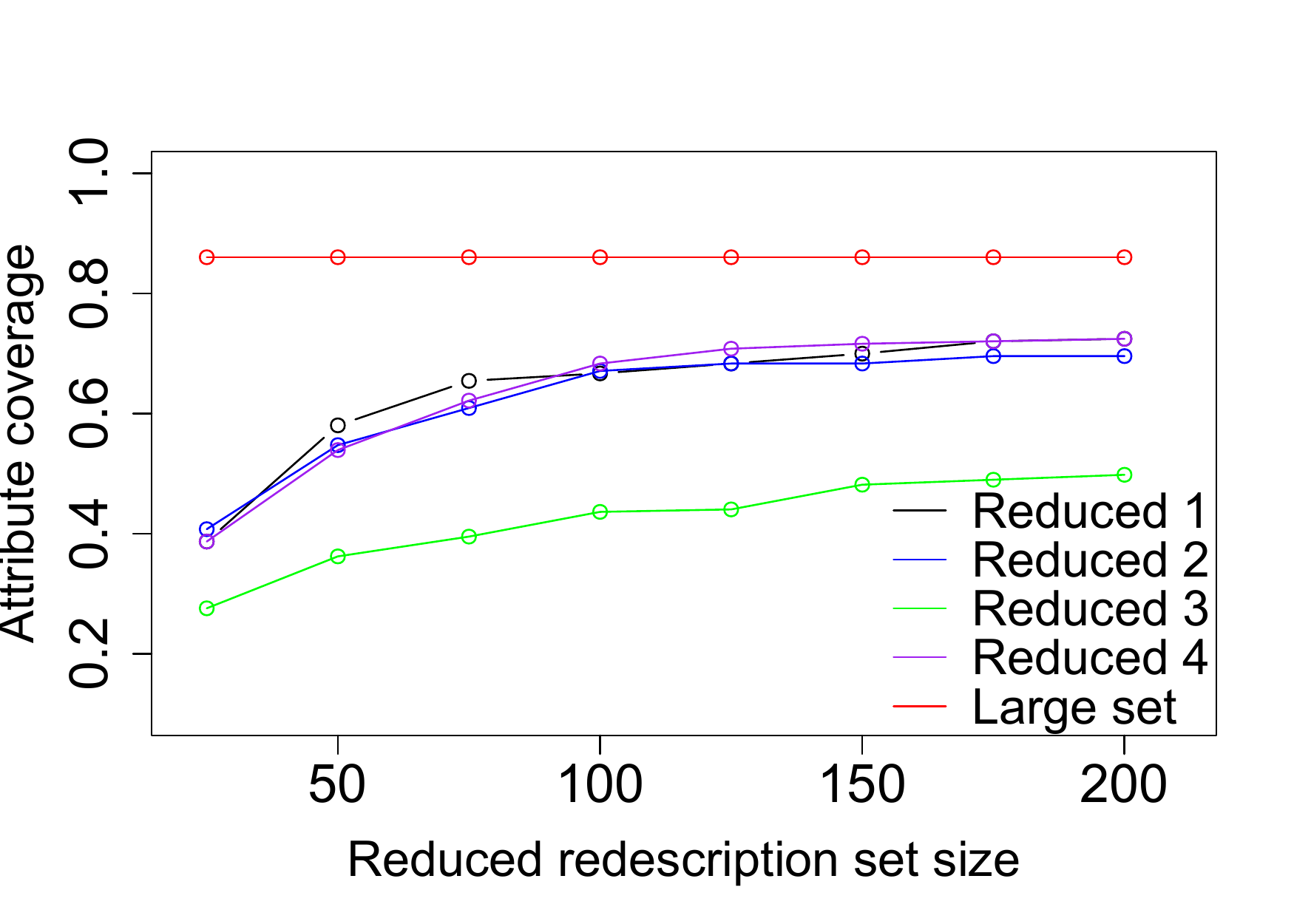}
  }\qquad
  \subfloat{%
  \includegraphics[width=0.42\textwidth]{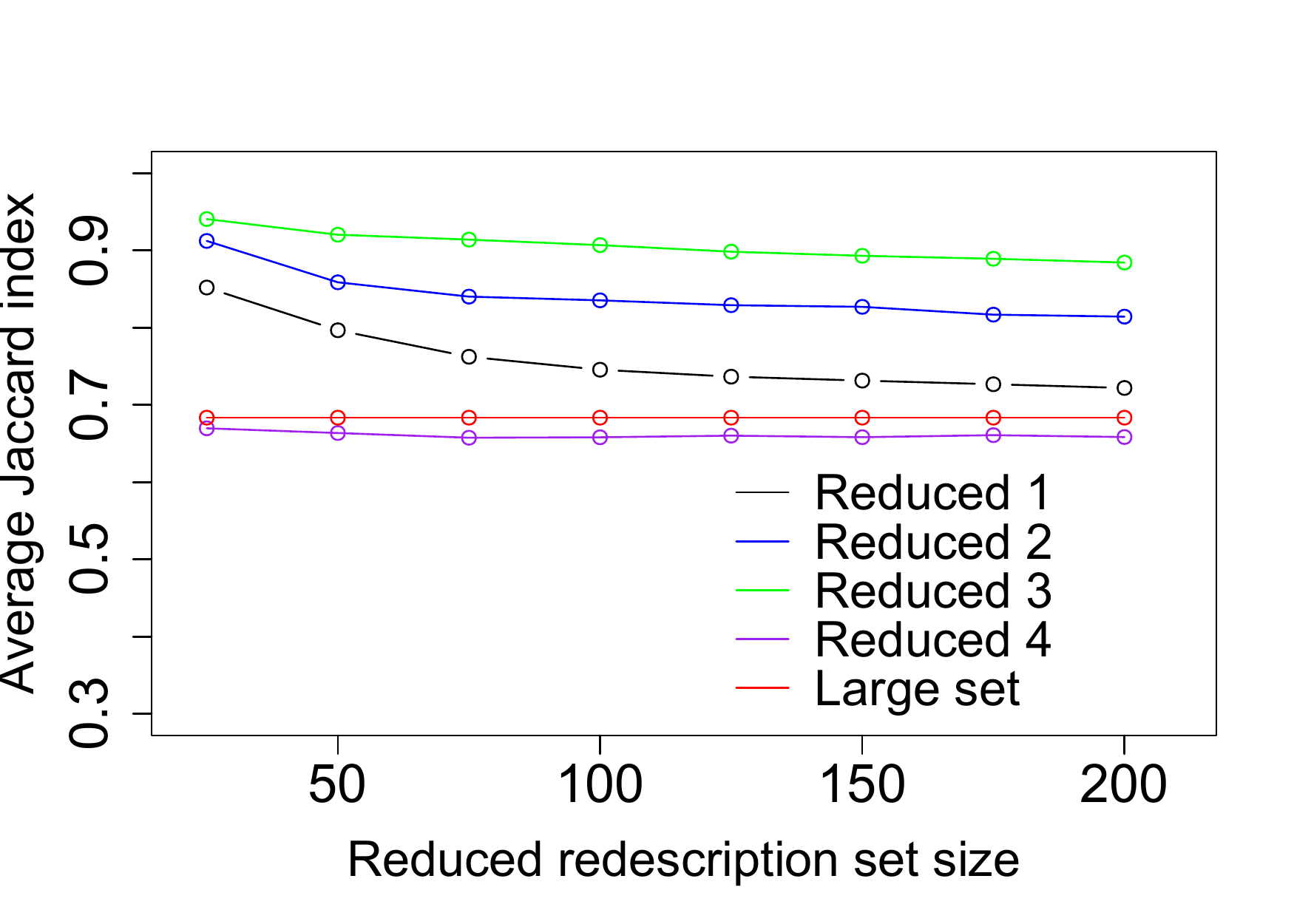}
  }\qquad
  \subfloat{%
    \includegraphics[width=0.42\textwidth]{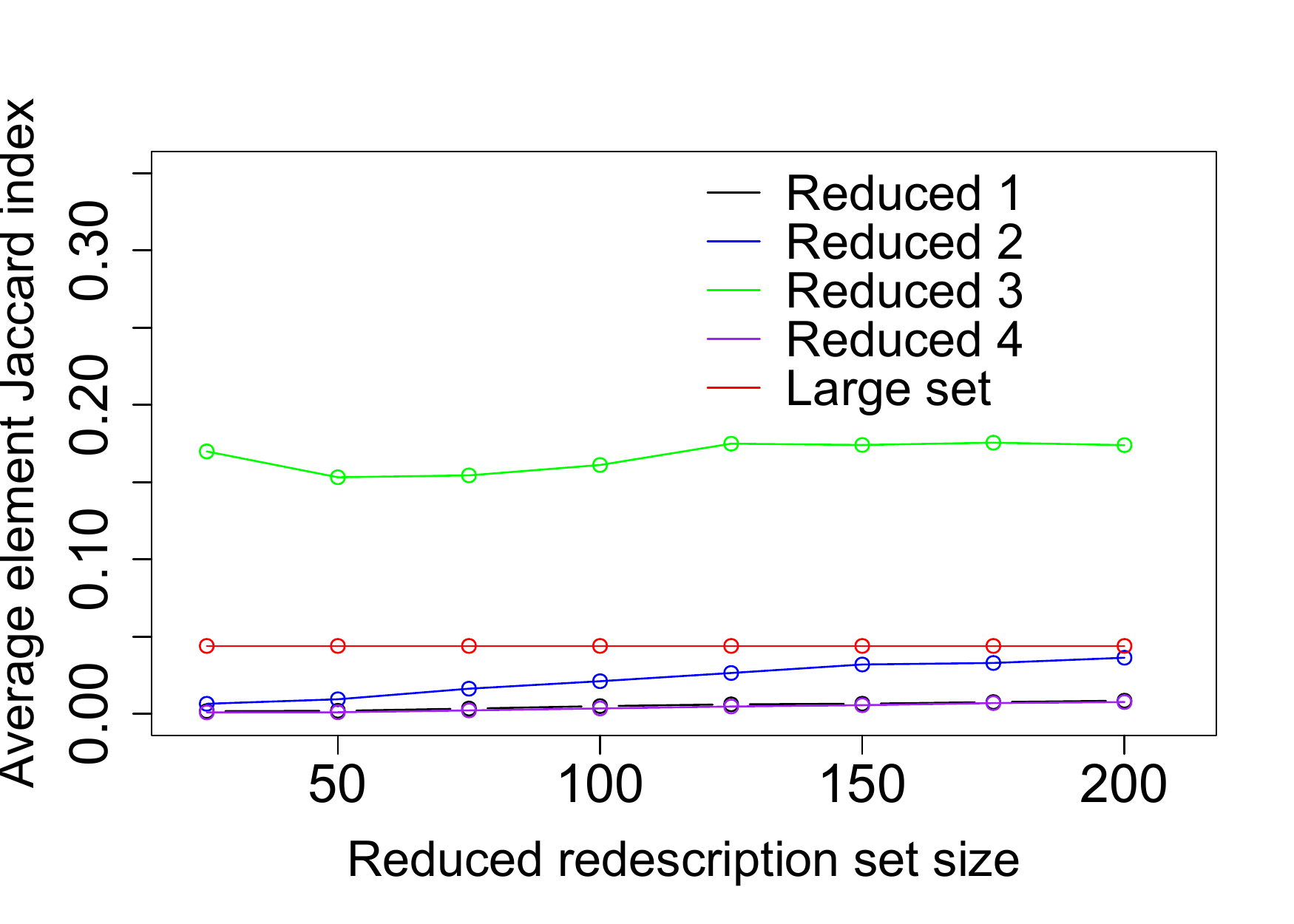}
  }\qquad
  \subfloat{%
  \includegraphics[width=0.42\textwidth]{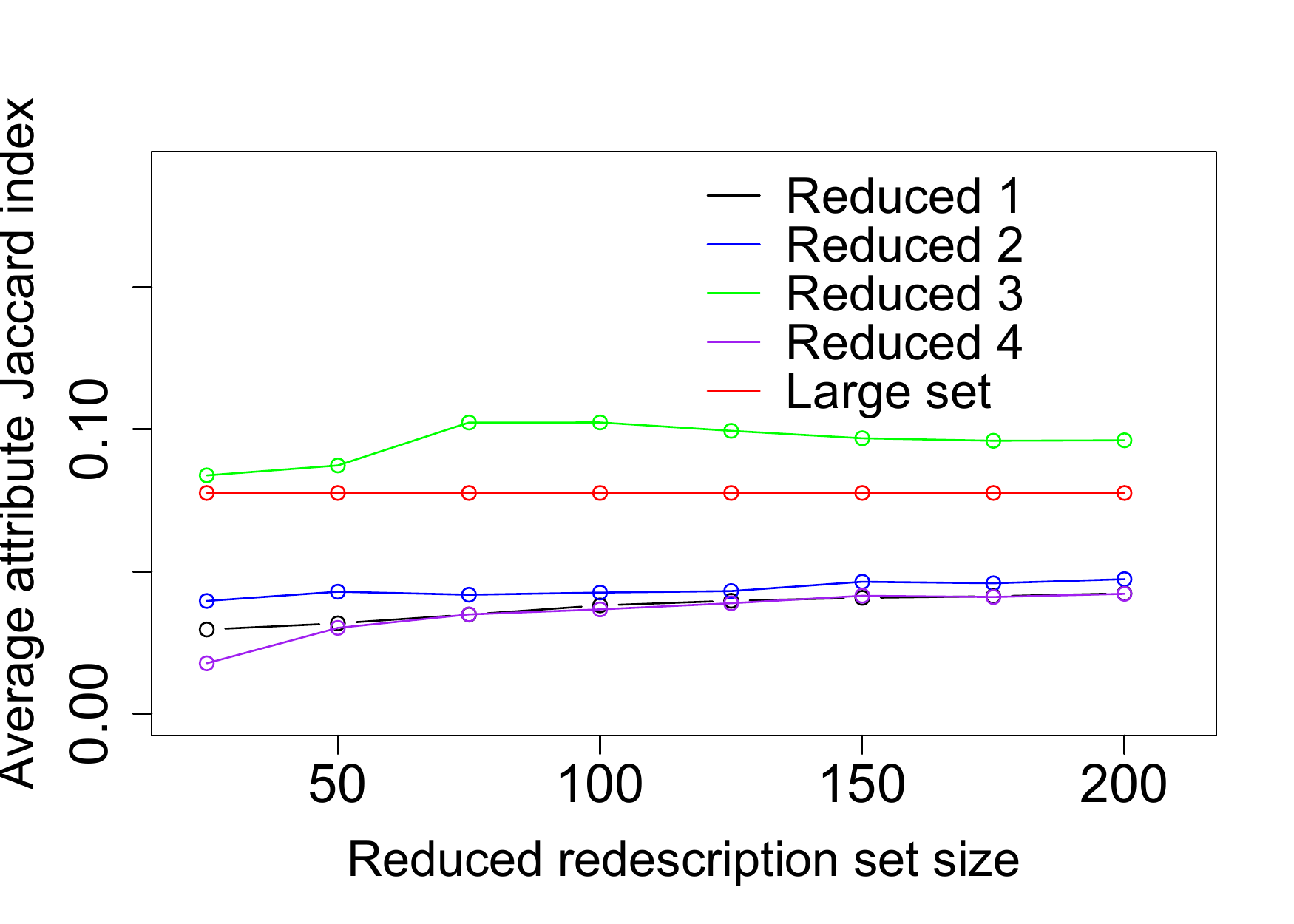}
}\qquad
\subfloat{%
    \includegraphics[width=0.42\textwidth]{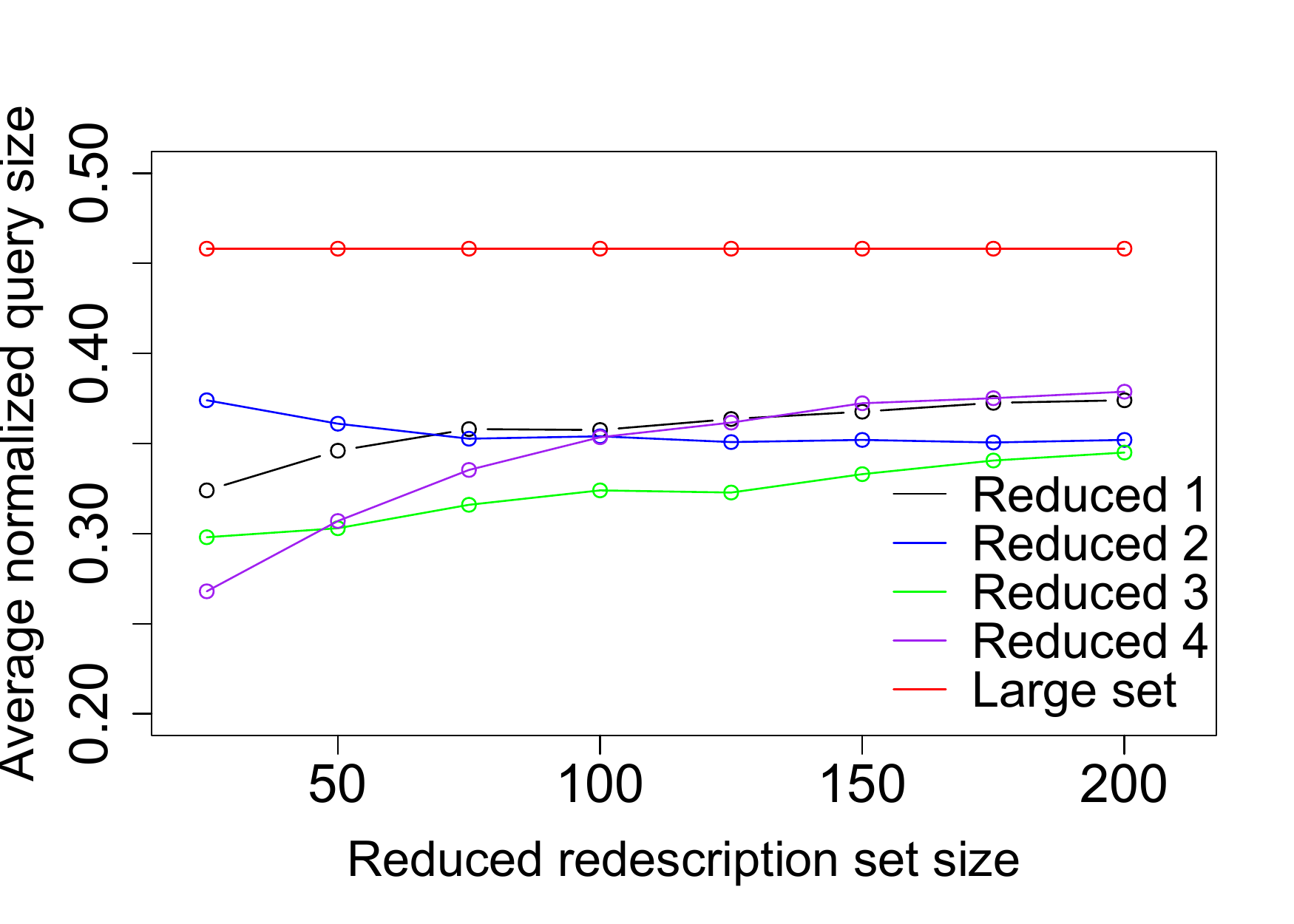}
}
\qquad
\subfloat{%
    \includegraphics[width=0.42\textwidth]{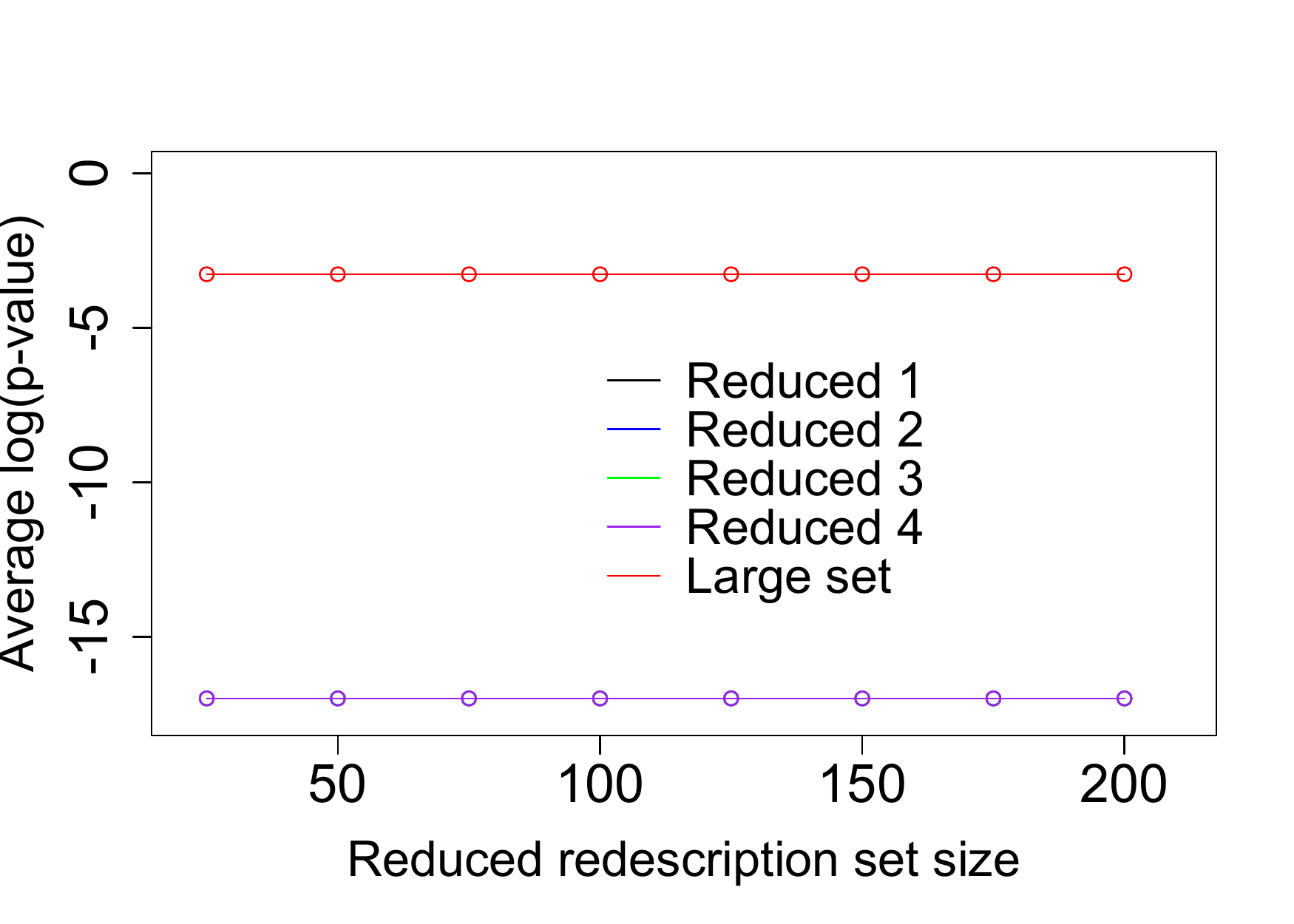}
    }
\qquad
\subfloat{%
    \includegraphics[width=0.42\textwidth]{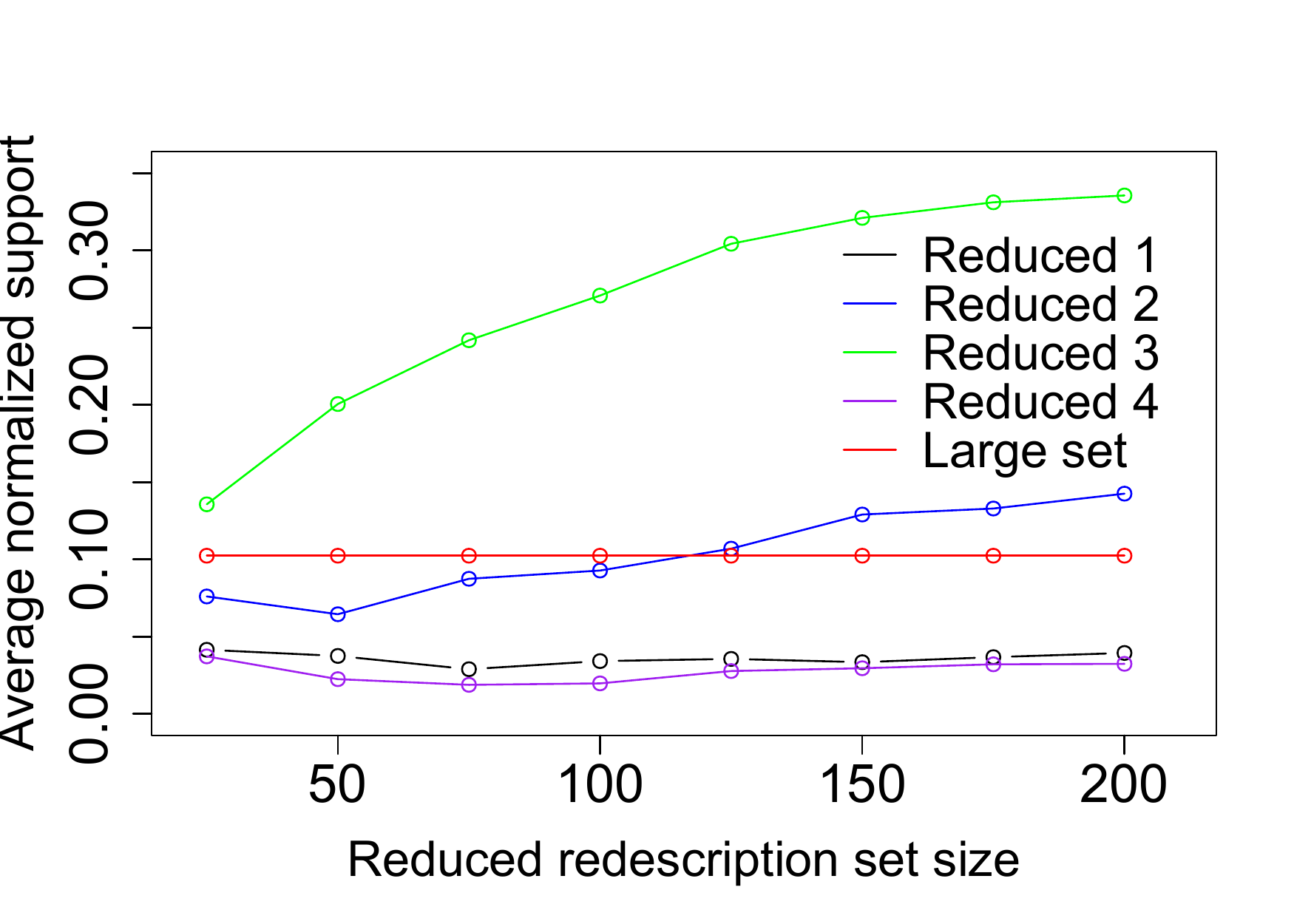}
}
  \caption{Plots comparing element and attribute coverage, average redescription: Jaccard index, log($p$-value), element/attribute Jaccard index, normalized support and normalized query size for resulting reduced sets of different size and the original, large redescription set containing all produced redescriptions. \emph{Reduced} $k$, corresponds to the reduced set obtained with the importance weights from the $k$-th row of the weight matrix $W$.}
   \label{fig:GRCPlots}
\end{figure*}

\noindent We start the analysis by examining the properties of the large redescription set presented in Figure \ref{fig:RSAnal}.
In Figure \ref{fig:GRCPlots}, we compare  the properties of redescriptions in the large redescription set, against properties of redescriptions in reduced sets based on different preference vectors. The results are presented only for the Bio dataset, however similar analysis for the DBLP and the Country dataset is presented in Section S2.2.3 (Online Resource 1).

Figure \ref{fig:RSAnal} shows distributions of quality measures for redescriptions in the large redescription set constructed with CLUS-RM algorithm. Redescription Jaccard index is mostly in $[0.6,0.7]$ interval, though a noticeable number is in $[0.9,1.0]$. The $p$-value is at most $0.01$ but mainly smaller than $10^{-17}$. The maximum average element Jaccard index equals $0.13$ and the maximum average attribute Jaccard index equals $0.14$ which shows a fair level of diversity among produced redescriptions. Over $99\%$ of redescriptions contain less than $15$ attributes in both queries, and more than $50\%$ contains less than $10$ attributes in both queries which is good for understandability. 

Plots in Figure \ref{fig:GRCPlots} contain $5$ graphs demonstrating a specific property of the reduced redescription set and its change with the increase of reduced redescription set size. The \emph{Reduced} $k$ graph demonstrates properties of redescriptions contained in redescription set created with the preference weights from the $k$-th row of $W$. The graph labelled \emph{Large set} demonstrates properties of redescriptions from a redescription set containing all produced redescriptions.

Increasing the importance weight for a redescription Jaccard index has the desired effect on redescription accuracy in the reduced sets of various size. Large weight on this criteria leads to sets with many highly accurate but more redundant redescriptions (average element Jaccard $>0.15$) with larger support (average support $>10\%$ of the total number of elements in the dataset). Consequence of larger support is increased overall element coverage. The effect is in part the consequence of using the Bio dataset that contains a number of accurate redescriptions with high support (also discussed in \citep{GalbrunPhd}). This effect is not observed on the Country and the DBLP dataset (Figures S4 and S5), where element and attribute coverage is increased only with increasing diversity weights in the preference vector. The average redescription Jaccard index decreases as the reduced set size increases which is expected since the total number of redescriptions with the highest possible accuracy is mostly smaller than $200$.

Use of weights from the second row of the importance matrix $W$ largely reduces redundancy and moderately lowers redescription accuracy in produced redescription set compared to weights that highly favour redescription accuracy. The equal weight combination provides accurate redescriptions (above large set average) that describe different subsets of elements by using different attributes (both below large set average). The average redescription support is lower as a result, around $5\%$ of data elements. Despite this, the element coverage is between $88\%$ and $100\%$ with the sharp increase to $98\%$ for a set containing $50$ redescriptions. The element coverage reaches $100\%$ for sets containing at least $175$ redescriptions.

Depending on the application, it might be interesting to find different, highly accurate descriptions of the same or very similar sets of elements (thus the weights from the third row of $W$ from Table \ref{tab:WMat} would be applied). Higher redundancy provides different characteristics that define the group. It sometimes also provides more specific information about subsets of elements of a given group.

We found several highly accurate redescriptions describing very similar subsets of locations on the Bio dataset by using weights from the third row of the matrix $W$. These locations are characterized as a co-habitat of the Arctic fox and one of several other animals with some specific climate conditions. We provide two redescriptions describing a co-habitat of the Arctic fox and the Wood mouse.

\par \noindent $q_1:\  -9.5 \leq t_{11}^- \leq 0.9 \ \wedge\ 9.7 \leq t_{7}^+ \leq 13.4$\par \noindent
$q_2:\ \text{Woodmouse } \ \wedge\ \text{ArcticFox} \ \wedge \ \neg \text{ MountainHare }$\par \noindent

\noindent This redescription describes $57$ locations with Jaccard index $0.83$. One very similar redescription describing $58$ locations from which $57$ are the same as above, with Jaccard index $0.87$ is: 

\par \noindent $q_1:\ -5.5 \leq \tilde{t}_{2} \leq 2.2 \ \wedge\ 6.4 \leq t_{9}^+ \leq 10.6$\par \noindent
$q_2:\  \text{Woodmouse } \ \wedge\ \text{ArcticFox} \ \wedge \ \neg \text{ Norwaylemming }$\par \noindent

Examples that are even more interesting can be found on the Country data where very similar sets of countries can be described by using different trading and general country properties. The example can be seen in Section S2.1.3, Figure S11 (Online Resource 1).

\subsubsection{Using the redescription variability index on the Country dataset}

\noindent We analyse the impact of missing values to redescription creation and use newly defined redescription variability index ($RW$), in the context of generalized set generation, on the Country dataset with a weight matrix shown in Table \ref{tab:WMat1}. The variability weight is gradually increased while other weights are equally decreased to keep the sum equal to $1.0$ (which is convenient for interpretation).

\begin{table}[ht!]
\caption{The weight matrix designed to explore the effects of changing redescription variability index on the resulting redescription set. These weights are applied on data containing missing values. Otherwise, the variability index weight (RV) should equal $0$.}
$$W_{miss}=\begin{bmatrix}
    \text{J}  & \text{pV} & \text{AJ}& \text{EJ} & RQS & RV \\
    0.2       & 0.2 & 0.2 & 0.2 & 0.19 & 0.01 \\
    0.18      & 0.18 & 0.18 & 0.18 & 0.18 & 0.1 \\
    0.14	  & 0.14 & 0.14 & 0.14 & 0.14 & 0.3 \\
    0.1       & 0.1 & 0.1 & 0.1 & 0.1 & 0.5 \\
    0.06 & 0.06 & 0.06 & 0.06 & 0.06 & 0.7 
\end{bmatrix}$$
\label{tab:WMat1}
\end{table}

\begin{figure}[ht]
    \centering
  \includegraphics[width=0.44\textwidth]{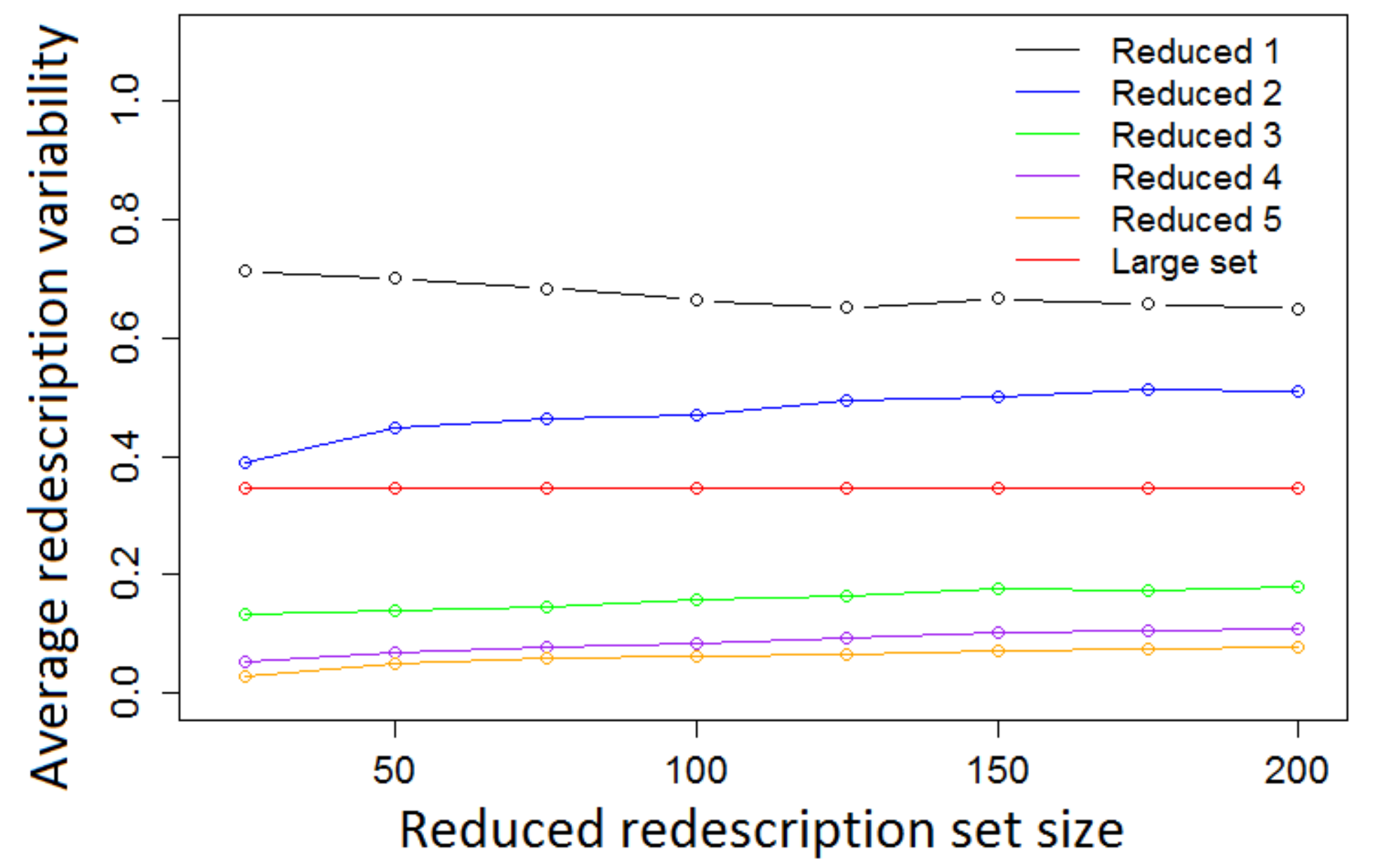}
  \caption{Change in average variability index of redescriptions in reduced redescription set for various set sizes and the set containing all created redescriptions.}
   \label{fig:Variability}
\end{figure}

The change in variability index depending on a reduced set size and comparison with the large set can be seen in Figure \ref{fig:Variability}.

\noindent As expected, increasing the importance weight for redescription variability favours selecting more stable redescriptions to the changes in missing values.

\begin{figure}[ht]
    \centering
  \includegraphics[width=0.44\textwidth]{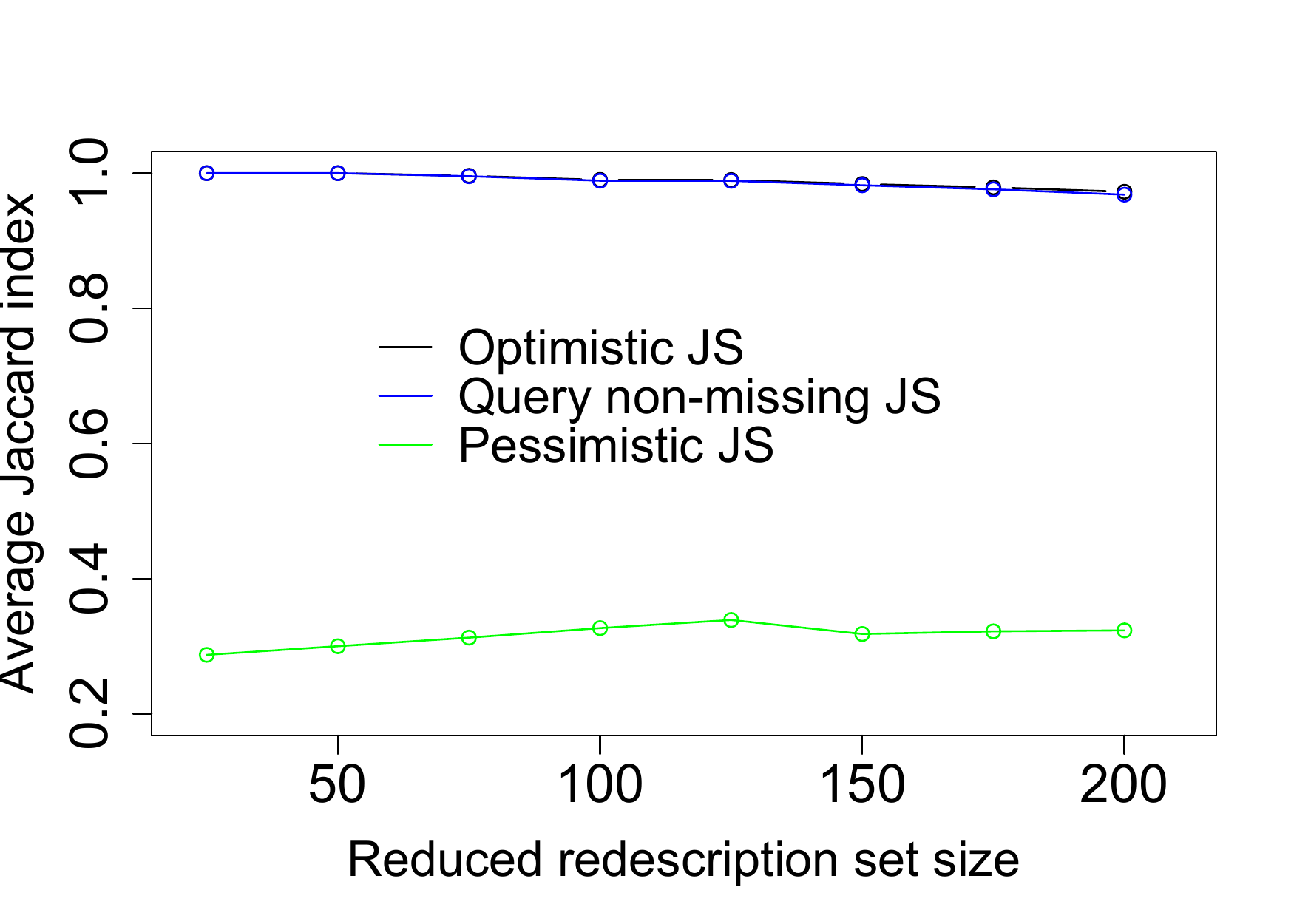}
  \caption{Optimistic, query non-missing and pessimistic Jaccard index for reduced sets of different sizes created with importance weight from the first row of the weight matrix $W_{miss}$.}
   \label{fig:VariabilityR1}
\end{figure}

\begin{figure}[ht]
    \centering
  \includegraphics[width=0.44\textwidth]{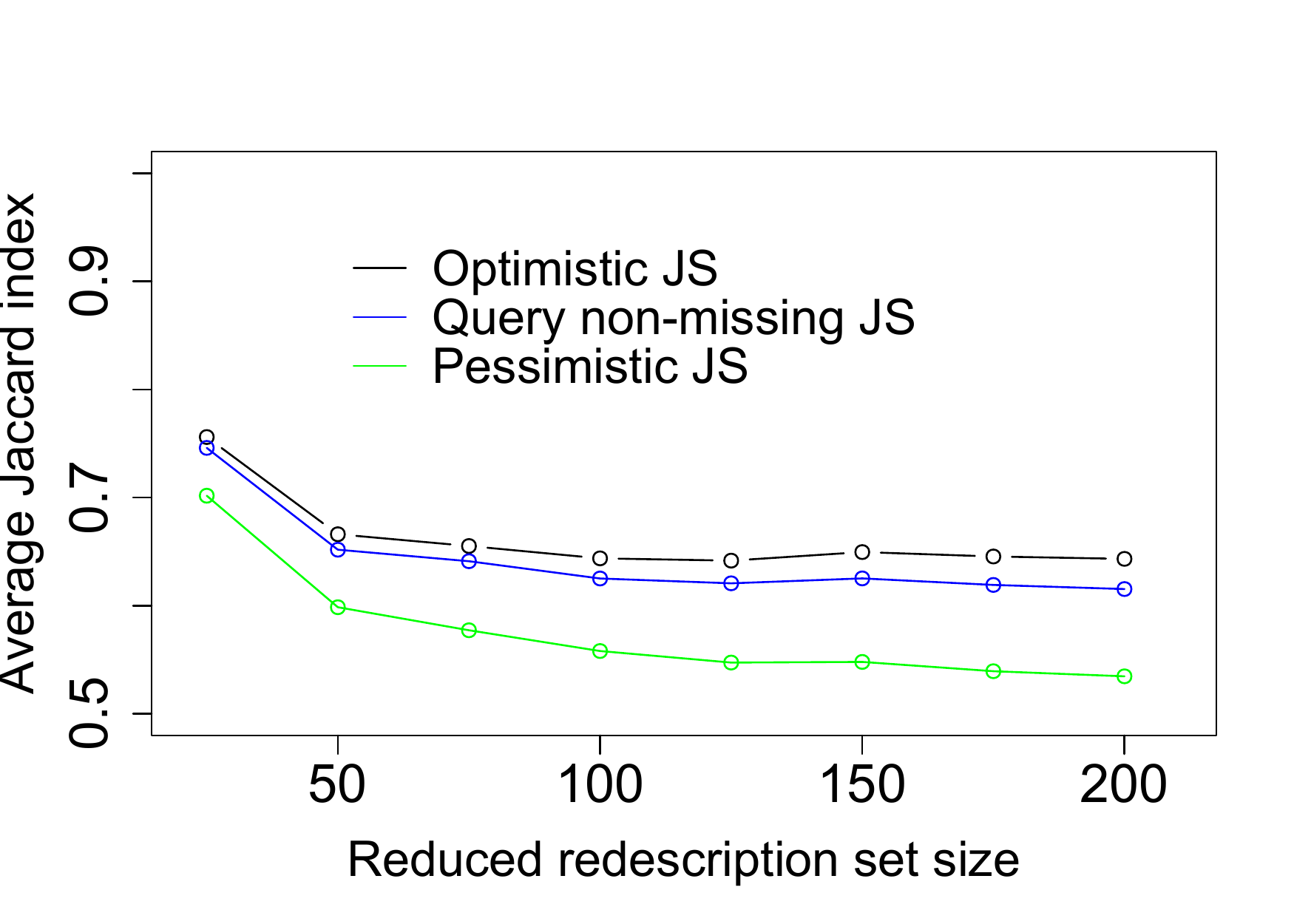}
  \caption{Optimistic, query non-missing and pessimistic Jaccard index for reduced sets of different sizes created with importance weight from the fourth row of the weight matrix $W_{miss}$.}
   \label{fig:VariabilityR4}
\end{figure}

To demonstrate the effects of variability index to redescription accuracy, we plot graphs comparing averages of optimistic, query non-missing and pessimistic Jaccard index for every row of the weight matrix for different reduced set sizes. The results for row $1$ and row $4$ can be seen in Figures \ref{fig:VariabilityR1} and \ref{fig:VariabilityR4}. Plots for reduced sets obtained with importance weights from the $2.$, the $3.$ and the $5.$ row of $W_{miss}$ are available in Figure S12 (Online resource 1).

\noindent Increasing the weight on the variability index has the desired effect of reducing the difference between values of different Jaccard index measures. However, the average optimistic and query non-missing Jaccard index values in the reduced sets drop as a result.

\noindent Redescription with $J_{qnm}=J_{pess}=J_{opt}= 1.0$:
\par \noindent $q_1:\ 3.6 \leq MORT \leq 4.1\ \wedge\  25.9 \leq \text{RUR\_POP} \leq 38.4$\par \noindent
$\ \wedge\  58.8 \leq \text{ LABOR\_PARTICIP\_RATE } \leq 61.1$ \par \noindent 
$q_2:\ 68.0 \leq E_{23} \leq 79.0\ \wedge\   0.7 \leq E/I_{104} \leq 4.4$\par \noindent
$\ \wedge\ 0.9 \leq E/I_{50}\leq 1.5 $

\noindent is highly accurate and stable redescription constructed by CRM-GRS with the importance weight from the fourth row of a matrix $W_{miss}$. It is statistically significant with the $p$-value smaller than $10^{-17}$. 

Redescriptions exist for which $J_{qnm}=J_{opt}$ and $J_ {pess}<J_{opt}$. In such cases, the drop in accuracy from $J_{opt}$ to $J_{pess}$ occurs because a number of elements exist in the dataset for which membership in the support of neither redescription query can be determined, due to missing values. Optimizing pessimistic Jaccard index is very strict and can discard some potentially significant redescriptions such as: 

\par \noindent $q_1:\ 5.6 \leq \text{ EMPL\_BAD } \leq 18.2\ \wedge\ 2.9 \leq MORT \leq 4.5 $ \par \noindent
$ \ \wedge\   2.0 \leq \text{ AGR\_EMP } \leq 10.5 \ \wedge\  -2.4 \leq \text{BAL} \leq 10.1$ \par \noindent
$q_2:\ 1.1 \leq E/I_{85} \leq 3.1 \ \wedge\ 93.0 \leq E_{97} \leq 98.0$. 
This redescription has $J_{qnm}=J_{opt}=1.0$ and $J_{pess}=0.48$. With the variability index of $0.52$ it describes all elements that can be evaluated by at least one redescription query with the highest possible accuracy.

This example motivates optimizing query non-missing Jaccard with positive weight on the variability index. It is especially useful when small number of highly accurate redescriptions can be found and when a large percentage of missing values is present in the data. 

\subsection{Evaluating the conjunctive refinement procedure}
\label{refEval}

\noindent The next step is to evaluate the conjunctive refinement procedure and its effects on the overall redescription accuracy. We use the same experimental set-up as in Section \ref{setQ} for both sets with the addition of the minimum refinement Jaccard index parameter, which was set to $0.4$ on the Bio dataset and $0.1$ on the Country and the DBLP dataset. The algorithm requires the initial clusters to start the mining process as explained in Section \ref{ruleredconstr} and in \citep{Mihelcic15LNAI}. To maintain the initial conditions, we create one set of initial clusters and use them to create redescriptions with and without the conjunctive refinement procedure. Since we use PCTs with the same initial random generator seed in both experiments, the differences between sets are the result of applying the conjunctive refinement procedure. The effects of using conjunctive refinement are examined on sets containing all redescriptions produced by CLUS-RM and on reduced sets created with equal importance weights by the generalized redescription set construction procedure (Row $1$ in matrix $W$).

\begin{figure}[ht]
    \centering
  \includegraphics[width=0.41\textwidth]{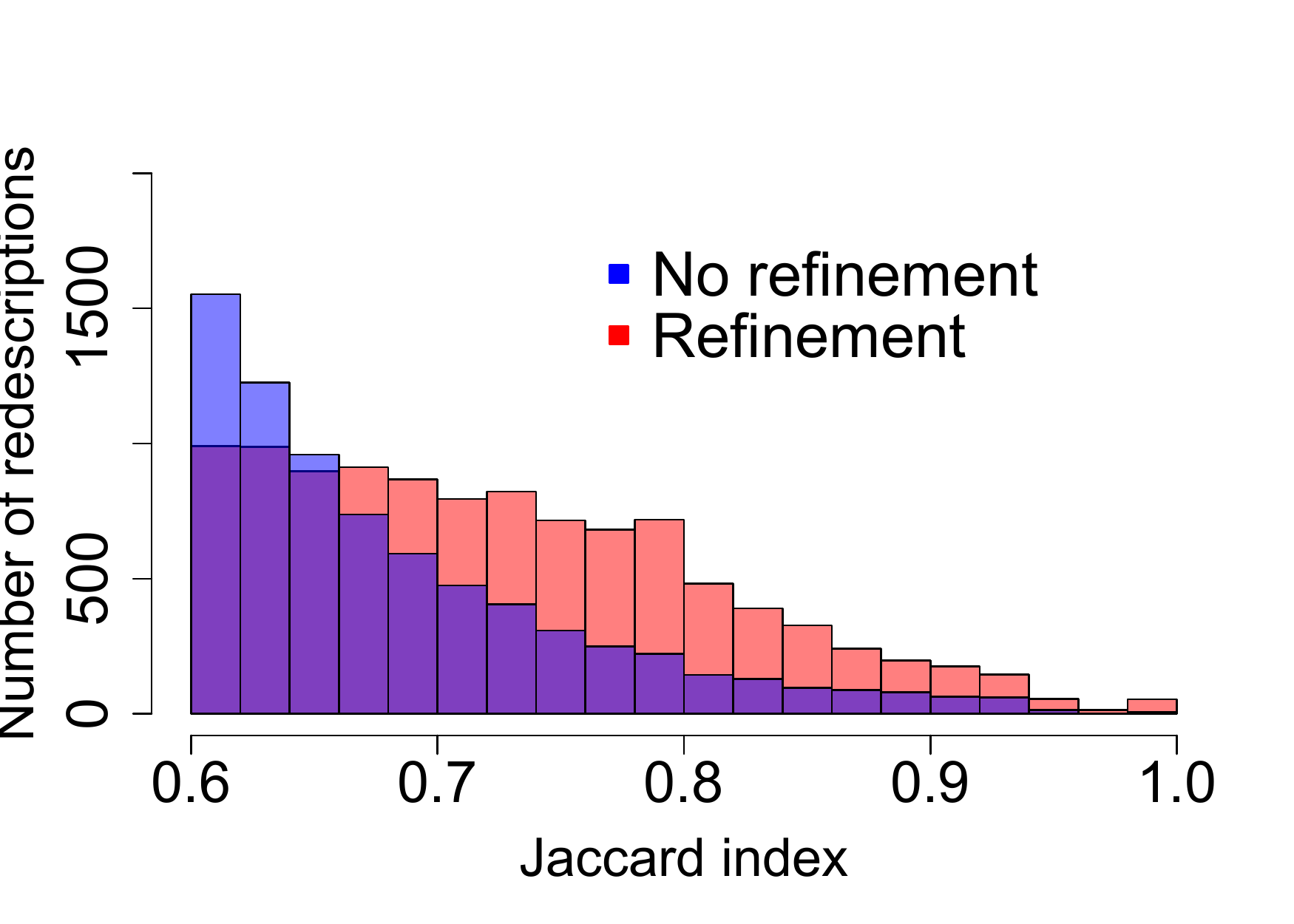}
  \caption{Distribution of a redescription Jaccard index in a large set created on a Bio dataset with and without the conjunctive refinement procedure. The set obtained without using the conjunctive refinement procedure contains $7413$ redescriptions, and the set obtained by using the conjunctive refinement procedure contains $10472$ redescriptions.}
   \label{fig:ComHistJNJ}
\end{figure}

The effects of using the refinement procedure on redescription accuracy are demonstrated in comparative histogram (Figure \ref{fig:ComHistJNJ}) showing the distribution of redescription Jaccard index in a set created by CLUS-RM with and without the refinement procedure. 

CLUS-RM produced $7413$ redescriptions, satisfying constraints from Section \ref{expproc}, without the refinement procedure and $10472$ redescriptions with the refinement procedure. The substantial increase in redescriptions satisfying user-defined constraints, when the conjunctive refinement procedure is used, is accompanied by significant improvement in redescription accuracy.

We performed the one-sided independent 2-group Mann-Whitney U test with the null hypothesis that there is a probability of $0.5$ that an arbitrary redescription ($R_{r}$) from a set obtained by using conjunctive refinement has the Jaccard index larger than the arbitrary redescription ($R_{nr}$) from a set obtained without using the conjunctive refinement procedure ($P(J(R_{r})>J(R_{nr}))=0.5$). The $p$-value of $2.2\cdot 10^{-16}$ lead us to reject the null hypothesis with the level of significance $0.01$ and conclude that $P(J(R_{r})>J(R_{nr}))> 0.5$ must be true.

\begin{figure}[ht]
    \centering
  \includegraphics[width=0.41\textwidth]{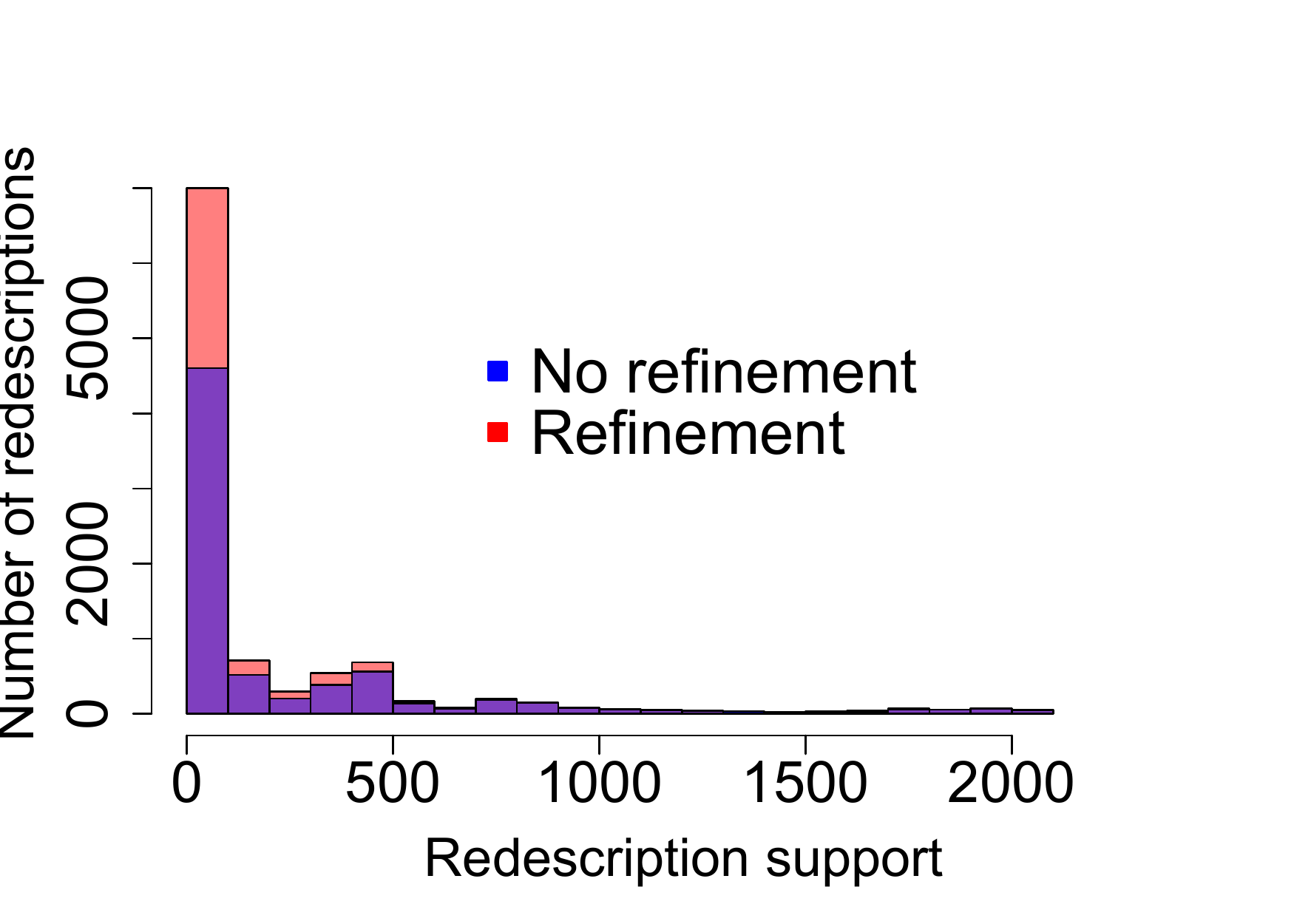}
  \caption{Distribution of a redescription support size in a large set created on a Bio dataset with and without the conjunctive refinement procedure. The set obtained without using the conjunctive refinement procedure contains $7413$ redescriptions, and the set obtained by using the conjunctive refinement procedure contains $10472$ redescriptions.}
   \label{fig:ComHistJNJSupp}
\end{figure}

Another useful property of the conjunctive refinement procedure is that it preserves the size of redescription support. The comparative distribution of redescription supports between the sets is shown in Figure \ref{fig:ComHistJNJSupp}.

\noindent Majority of $3059$ redescriptions that entered the redescription set because of the improvements made by the conjunctive refinement have supports in the interval $[10,500]$ elements. Because of that, the average support size in the redescription set obtained by using the refinement procedure ($217.98$) is lower than that obtained without the refinement procedure ($263.63$). The change in distribution is significant, as shown by the one-sided independent 2-group Mann-Whitney U test. The test rejects the hypothesis $P(|supp(R_{nr})|>|supp(R_{r})|)=0.5$ with the level of significance $0.01$ ($p$-value equals $2.4\cdot 10^{-14}$), thus showing that $P(|supp(R_{nr})|>|supp(R_{r})|) > 0.5$. 

Using the conjunctive refinement procedure improves redescription accuracy and adds many new redescriptions to the redescription set. However, since the reduced sets are presented to the user, it is important to see if higher quality reduced sets can be created from the large set by using the conjunctive refinement procedure compared to the set obtained without using the procedure.

\begin{figure}[ht!]
\includegraphics[width=0.48\textwidth]{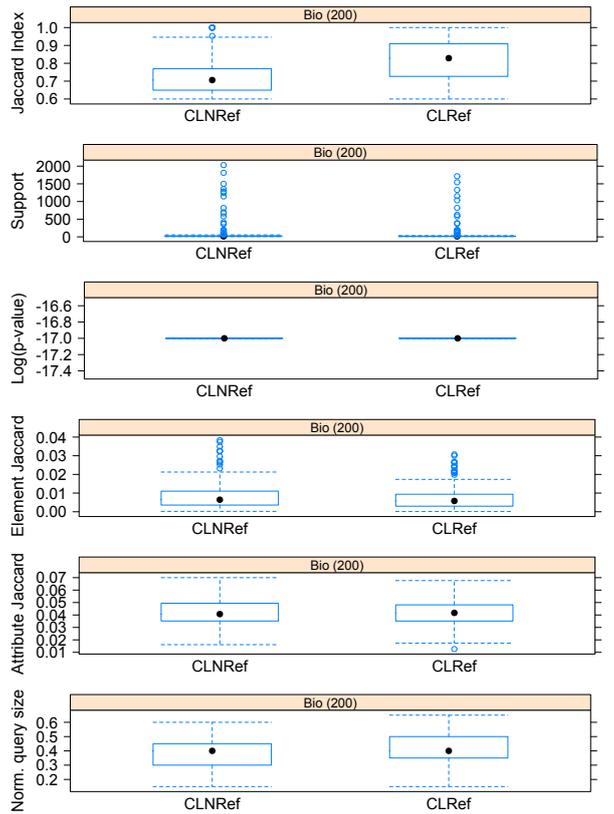}
\caption{Boxplots comparing distributions of redescription: Jaccard index, support, log($p$-value), element Jaccard index, attribute Jaccard index and normalized query size in reduced sets containing $200$ redescriptions. The reduced sets were obtained by the generalized redescription set construction procedure by using equal importance weight for each measure.}
   \label{fig:CompBPBio}
\end{figure}

We plot comparative distributions for all defined redescription measures for reduced sets extracted from the redescription set obtained with (\emph{CLRef}) and without (\emph{CLNRef}) the conjunctive refinement procedure. The comparison made on the sets containing $200$ redescriptions is presented in Figure \ref{fig:CompBPBio}. The boxplots representing distributions of supports show that the redescription construction procedure extracts redescriptions of various support sizes, which was intended to prevent focusing only on large or small redescriptions based on redescription accuracy.

We compute the one-sided independent 2-group Mann-Whitney U test on the reduced sets for the redescription Jaccard index ($J$) and the normalized redescription query size ($RQS$) since there seem to be a difference in distributions as observed from Figure \ref{fig:CompBPBio}. For other measures, we compute the two-sided Mann-Whitney U test to assess if there is any notable difference in values between the sets.
 
The null hypothesis that $P(J(R_{r})>J(R_{nr}))=0.5$ is rejected with the $p$-value smaller than $2.2\cdot 10^{-16}<0.01$, thus the alternative hypothesis $P(J(R_{r})>J(R_{nr})) > 0.5$  holds. The difference in support between two sets is not statistically significant ($p$-value equals $0.21$, obtained with the two-sided test). Distributions of redescription $p$-values are identical because all redescriptions have equal $p$-value: $0.0$. The difference in average attribute/element Jaccard index is also not statistically significant ($p$-values $0.88$ and $0.13$ respectively obtained with the two-sided test). The $p$-value for the null hypothesis $P(RQS(R_{nr})<RQS(R_{r})) = 0.5$ equals $5.25\cdot 10^{-6}<0.01$ thus the alternative hypothesis $P(RQS(R_{nr})<RQS(R_{r})) > 0.5$ holds. 

The refinement procedure enables constructing reduced sets containing more accurate redescriptions with the average Jaccard index increasing from $0.72$, for reduced set obtained without using refinement procedure, to $0.82$ for reduced set obtained when refinement procedure is used. This improvement sometimes increases redescription complexity, albeit this is limited on average to having less than $1$ additional attribute in redescription queries.

The set produced by using the conjunctive refinement procedure has the element coverage of $0.9996$ and the attribute coverage of $0.7613$ compared to the set where this procedure was not used where the element coverage is $1.0$ and the attribute coverage is $0.7243$. 

The conjunctive refinement procedure also significantly increases redescription accuracy on the DBLP and the Country dataset. Equivalent analysis for these datasets is performed in Section S2.3 (Online Resource 1).

\subsection{Comparisons with other state of the art redescription mining algorithms.}
\label{comparison}

\noindent In this section, we present the comparative results of redescription set quality produced by our framework (CRM-GRS) compared to the state of the art algorithms: the ReReMi \cite{GalbrunBW}, the Split trees and the Layered trees \cite{Zinchenko}. To perform the experiments, we used the implementation of the ReReMi, the Split trees and the Layered trees algorithm within the tool Siren \citep{GalbrunSiren}. 

The ReReMi algorithm was already compared in \citep{GalbrunBW} with the CartWheels algorithm \citep{RamakrishnanCart} (on a smaller version of a DBLP and the Bio dataset), with the association rule mining approach obtained by the ECLAT frequent itemset miner \citep{ZakiEclat} and the greedy approach developed by \cite{Gallo}.
The approach from \cite{ZakiReas}, which is also related, works only with boolean attributes and have no built in mechanism to differentiate different views. 
Redescription mining on the DBLP dataset with the original implementation of the algorithm\footnote{\url{http://www.cs.rpi.edu/~zaki/www-new/pmwiki.php/Software/Software}} returned $49$ redescriptions, however they only describe authors by using co-authorship network. Since, our goal is to describe authors by their co-authorship network and provide the information about the conferences they have published in, these redescriptions are not used in our evaluation. To use the approach on the Bio dataset, we first applied the \emph{Discretize} filter in weka\footnote{\url{http://www.cs.waikato.ac.nz/ml/weka/}} to obtain nominal attributes. Then, we applied \emph{NominalToBinary} filter to obtain binary attributes that can be used in Charm-L. As a result, the number of attributes on the Bio dataset increased to $1679$ making the process of constructing a lattice of closed itemsets to demanding with respect to execution time constraints. The Country dataset contains missing values which are not supported by this approach.

Since there is an inherent difference in the number of created redescriptions, depending on the type of logical operators used to create them, between CLUS-RM and the comparative algorithms, we split the algorithm comparison in two parts. First, we compare redescription properties created by using all logical operators and then redescriptions created by using only the conjunction and the negation operator (Bio and DBLP dataset) or only by using the conjunction operator (Country dataset). 

After obtaining redescriptions with the algorithms implemented in the tool Siren \citep{GalbrunSiren}, with parameters specified in Section \ref{expproc}, we used the \emph{Filter redundant redescriptions} option to remove duplicate and redundant redescriptions. Since SplitTrees and LayeredTrees algorithms always use all logical operators to create redescriptions, we created a redescription set with these approaches and filtered out redescriptions containing the disjunction operator in at least one of its queries. 

For each obtained redescription set from the ReReMi, the Split trees and the Layered trees algorithm, we extracted a redescription set of the same size with the generalized redescription set procedure with equal weight importance for each redescription criteria. These sets are extracted from a large set created with the CLUS-RM algorithm with the parameters specified in Section \ref{expproc}. 

We plot pairwise comparison boxplots for each redescription measure comparing the performance of our framework with the three chosen approaches.

For each comparison we analyse the hypothesis about the distributions by using the one-sided independent 2-group Mann-Whitney U test (see summary in Table \ref{table1}). 

\begin{table*}[htbp]
\centering
\caption{Table containing p-values obtained with the one-sided independent 2-group Mann-Whitney U test. We test the hypothesis to have the probability $0.5$ that the redescription chosen from the redescription set obtained by our framework has larger/smaller value compared to the redescription chosen from the redescription set produced by the ReReMi, the Split trees (ST) or the Layered trees (LT), depending on the redescription measure used, compared to the alternative in which a redescription chosen from a set produced by our framework has the probability greater than $0.5$ for this outcome. For the Jaccard index (J) and support we test if the probability is greater than $0.5$ to obtain larger values, for the average redescription redundancy based on elements/attributes contained in their support (AEJ)/ (AAJ) and redescription query size (RQS), we test if the probability is larger to obtain smaller values in the set produced by our framework. Each table cell contains two $p$-values in the format $pVal1$/$pVal2$. The first p-value relates to the set produced by the CLUS-RM without the conjunctive refinement procedure and the second with the refinement procedure.}
\begin{tabular}{|c|c|c|c|c|c|}
\hline
Dataset & Operators & Measure & ReReMi & ST & LT  \\ \hline
\multirow{ 12}{*}{Bio} & \multirow{ 6}{*}{AllOp (DCN)} & J & $0.91$/$2\cdot 10^{-4}$ \Tstrut\Bstrut & $2.6\cdot 10^{-9}$/$2.7\cdot 10^{-15}$ & $0.0035$/$1.9\cdot 10^{-7}$ \\
& & Supp & $1.0$/$1.0$ & $1.0$/$1.0$ & $0.9994$/$1.0$  \\
& & p-value & $2\cdot 10^{-9}$/$2\cdot 10^{-9}$ & $0.0217$/$0.0217$ & $0.0408/0.0408$  \\
& & AEJ & $<2\cdot 10^{-16}$/$<2\cdot 10^{-16}$ & $5.3\cdot 10^{-10}$/$3.4\cdot 10^{-11}$ & $1.3\cdot 10^{-5}$/$2.3\cdot 10^{-8}$   \\
& & AAJ & $2\cdot 10^{-7}$/$2\cdot 10^{-7}$  & $1.2\cdot 10^{-13}$/$<2\cdot 10^{-16}$ & $0.1122$/$8.2\cdot 10^{-5}$  \\
& & RQS & $2\cdot 10^{-8}$/$9\cdot 10^{-5}$ & $1.5\cdot 10^{-8}$/$1.3 \cdot 10^{-5}$ & $6.7\cdot 10^{-7}$/$5\cdot 10 ^{-5}$   \\ \cline{2-6}
& \multirow{ 6}{*}{ConjNeg (CN)} & J & $0.0035$/$1.5\cdot 10^{-12}$ \Tstrut\Bstrut &  &    \\ 
& & Supp & $1.0$/$1.0$ &  &    \\
& & p-value & $0.08$/$0.08$ &  &    \\
& & AEJ & $<2\cdot 10^{-16}$/$1.4\cdot 10^{-15}$ & $|\mathcal{R}|<10$ & $|\mathcal{R}|<10$    \\
& & AAJ & $<2\cdot 10^{-16}$/$<2\cdot 10^{-16}$ &  &    \\
& & RQS & $4.3\cdot 10^{-10}$/$3.5\cdot 10^{-7}$ &  &    \\ \hline
\multirow{ 12}{*}{DBLP} & \multirow{ 6}{*}{AllOp (DCN)} & J & $1.0$/$1.0$  \Tstrut\Bstrut & $1.0$/$0.9999$ &   \\
& & Supp & $1.0$/$1.0$ & $0.0033$/$0.0033$ &   \\
& & p-value & $1.0$/$1.0$ & $1.0$/$1.0$ &    \\
& & AEJ & $1.0$/$1.0$ & $0.904$/$0.980$ & $|\mathcal{R}|<10$   \\
& & AAJ & $1.0$/$1.0$ & $0.9997$/$0.9998$  &   \\
& & RQS & $<2\cdot 10^{-16}$/$8.6\cdot 10^{-9}$ & $<2\cdot 10^{-16}$/$3.5\cdot 10^{-15}$ &    \\ \cline{2-6}
& \multirow{ 6}{*}{ConjNeg (CN)} & J & $0.0127$/$5.96\cdot 10^{-7}$ &  &   \Tstrut\Bstrut \\ 
& & Supp & $1.74\cdot 10^{-8}$/$1.14\cdot 10^{-9}$ &  &   \\ 
& & p-value & $0.9779$/$0.9933$ &  &    \\
& & AEJ & $1.0$/$1.0$ & $|\mathcal{R}|<10$ & $|\mathcal{R}|<10$   \\
& & AAJ & $1.0$/$1.0$ &  &    \\
& & RQS & $1.0$/$1.0$ &  &   \\ \hline
\multirow{ 12}{*}{Country} & \multirow{ 6}{*}{AllOp (DCN)} & $J_{pess}$ & $1.0$/$0.9979$ \Tstrut\Bstrut&  &   \\
& & $J_{qnm}$ & $<2\cdot 10^{-16}$/$<2\cdot 10^{-16}$ &  &   \\
& & Supp & $1.0$/$1.0$ &  &   \\
& & p-value & $6.3\cdot 10^{-10}$/$7.5\cdot 10^{-10}$ & NA & NA    \\
& & AEJ & $<2\cdot 10^{-16}$/$<2\cdot 10^{-16}$ & NA & NA   \\
& & AAJ & $<2\cdot 10^{-16}$/$<2\cdot 10^{-16}$ &  &    \\
& & RQS & $<2\cdot 10^{-16}$/$<2\cdot 10^{-16}$ &  &   \\ \cline{2-6}
& \multirow{ 6}{*}{Conj (CN)} & $J_{pess}$ & $0.257$/$7\cdot 10^{-6}$ \Tstrut\Bstrut &  &    \\ 
& & $J_{qnm}$ & $5.2\cdot 10^{-7}$/$2.3\cdot 10^{-8}$ &  &    \\ 
& & Supp & $4.7\cdot 10^{-4}$/$0.769$ &  &    \\ 
& & p-value & $0.0503$/$0.0239$ & NA  & NA    \\
& & AEJ & $0.608$/$2.6\cdot 10^{ -5}$ & NA & NA   \\
& & AAJ & $1.74\cdot 10^{-15}$/$3.3\cdot 10^{-12}$ &   &    \\
& & RQS & $1.3\cdot 10^{-9}$/$3.7\cdot 10^{-17}$ &  &    \\ \hline

\end{tabular}
\label{table1}
\end{table*}

\subsubsection{Comparison on the Bio dataset}

\begin{figure}[ht]
    \centering

  \includegraphics[width=0.5\textwidth]{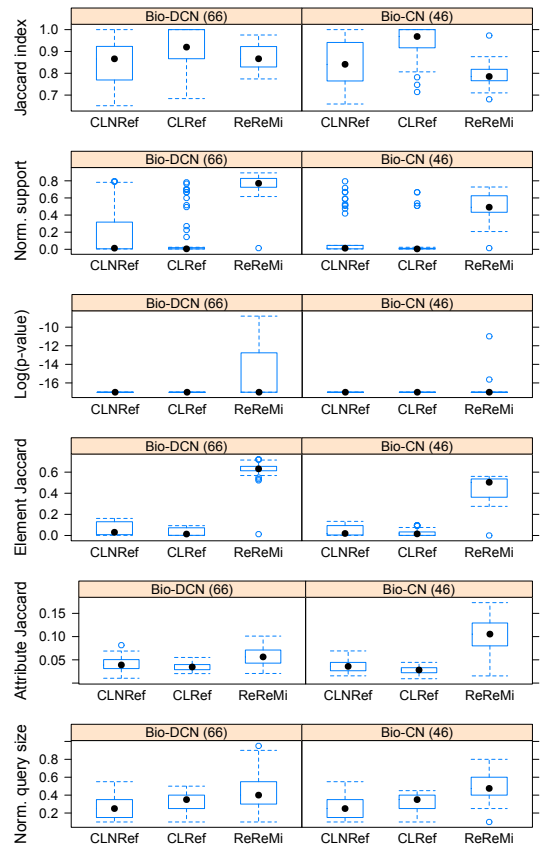}
\caption{Boxplots comparing redescriptions produced with our framework (CLNref, CLRef) and the ReReMi algorithm (ReReMi) on the Bio dataset. Sets contain $66$ redescriptions created by using all defined logical operators and $46$ redescriptions when only conjunction and negation operators are used to construct redescription queries.}
   \label{fig:RRBio}
\end{figure}

\begin{figure}[ht]
    \centering

  \includegraphics[width=0.5\textwidth]{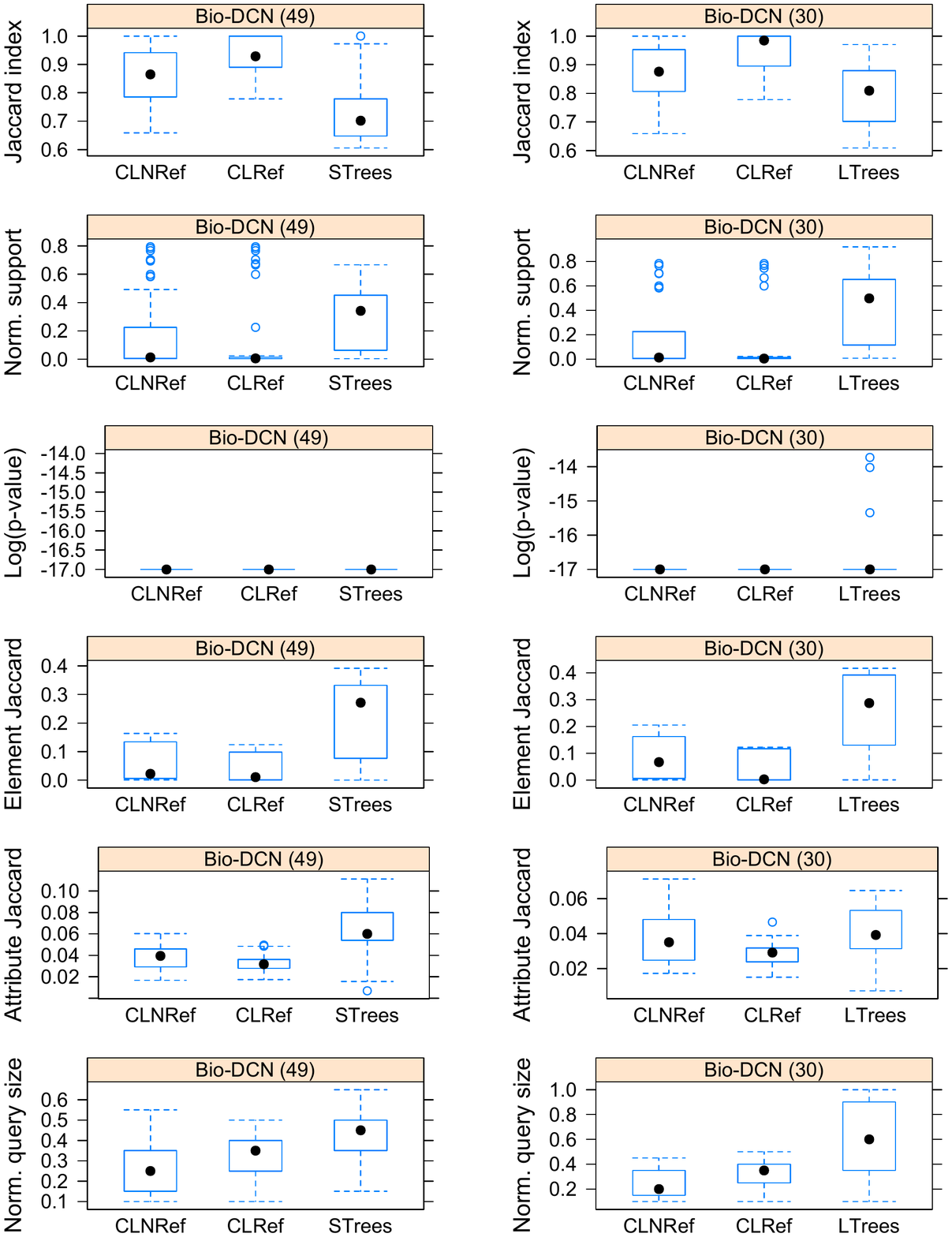}
\caption{Boxplots comparing $49$ redescriptions by using all defined logical operators, produced with our framework (CLNref, CLRef) and the Split trees algorithm (STrees) on the Bio dataset (left). The analogous comparison is made with the Layered trees algorithm (LTrees) on $30$ redescriptions (right).}
   \label{fig:STLTBio}
\end{figure}

\noindent First, we compare the algorithms on the Bio dataset. Figures \ref{fig:RRBio}, \ref{fig:STLTBio} and Table \ref{table1} show that the set produced by CRM-GRS tend to contain more accurate redescriptions on the Bio dataset when the conjunction and the negation operators are allowed and when the conjunctive refinement procedure is used compared to all other approaches.

\noindent The results are significant at the significance level of $0.01$, except for the case of ReReMi when all logical operators were allowed and refinement procedure was not used in the CLUS-RM algorithm. 
Redescriptions contained in redescription sets produced by CRM-GRS tend to have smaller $p$-values compared to redescriptions produced by other tree - based algorithms (statistically significant with the significance level of $0.05$). Redescription sets created by CRM-GRS tend to contain redescriptions with smaller element/attribute Jaccard index (redundancy) and smaller query size (the difference is statistically significant with the significance level of $0.01$ with the exception of a set created by CRM-GRS, when conjunctive refinement procedure was not used in CLUS-RM, compared to the set created by Layered trees algorithm). 

Element and attribute coverage analysis for all approaches is provided in Section S2.5.1 (Online Resource 1). This analysis suggests that despite smaller average redescription support, our framework has comparable performance with respect to element and attribute coverage.

As already discussed in \citep{GalbrunPhd}, the ReReMi algorithm has a drift towards redescriptions with large supports on the Bio dataset. The consequence is a large element redundancy among produced redescriptions. The Split trees and the Layered trees algorithms produce redescriptions in the whole support range, though majority of produced redescriptions still have a very high support resulting in large element redundancy. Our approach returns redescriptions with various support size as can be seen from Figures \ref{fig:RRBio} and \ref{fig:STLTBio} though majority of produced redescriptions are very close to the minimal allowed support. However, if needed, the minimal support can be adjusted to produce sets containing redescriptions that describe larger sets of elements. It is also possible to produce multiple sets, each being produced with different minimal and maximal support bounds. Also, by adjusting the importance weights to highly favour Jaccard index, the user can produce reduced sets with similar properties as those produced by the ReReMi, the Layered trees and the Split trees. The distribution of support size in the large redescription set produced with the basic variant of CLUS-RM algorithm on the Bio dataset can be seen in Figure \ref{fig:RSAnal}. The increase in accuracy obtainable by using different weights to construct reduced sets can be seen in Figure \ref{fig:GRCPlots}.

Redescription sets produced with the Layered and the Split trees algorithms do not create enough redescriptions containing only conjunction and negation operator in its queries to make the distribution analysis. The Layered trees algorithm produced only one redescription with Jaccard index $0.62$ and the Split trees algorithm created four redescriptions with Jaccard index $0.97$, $0.65$, $0.7$ and $0.78$. On the other hand, the CLUS-RM with the conjunctive refinement procedure created over $14000$ redescriptions containing only conjunction and negation in the queries with the Jaccard index greater than $0.6$ from which $73$ redescriptions have Jaccard index $1.0$.  

Our framework complements the existing approaches which is visible from redescription examples found by our approach that were not discovered by other algorithms. Section S2.5.1 (Online Resource 1) contains one example of very similar redescription, found by the ReReMi and the CRM-GRS, and several redescriptions discovered by CRM-GRS that were not found by other approaches.

\subsubsection{Comparison on the DBLP dataset}
\begin{figure}[ht]
    \centering

  \includegraphics[width=0.5\textwidth]{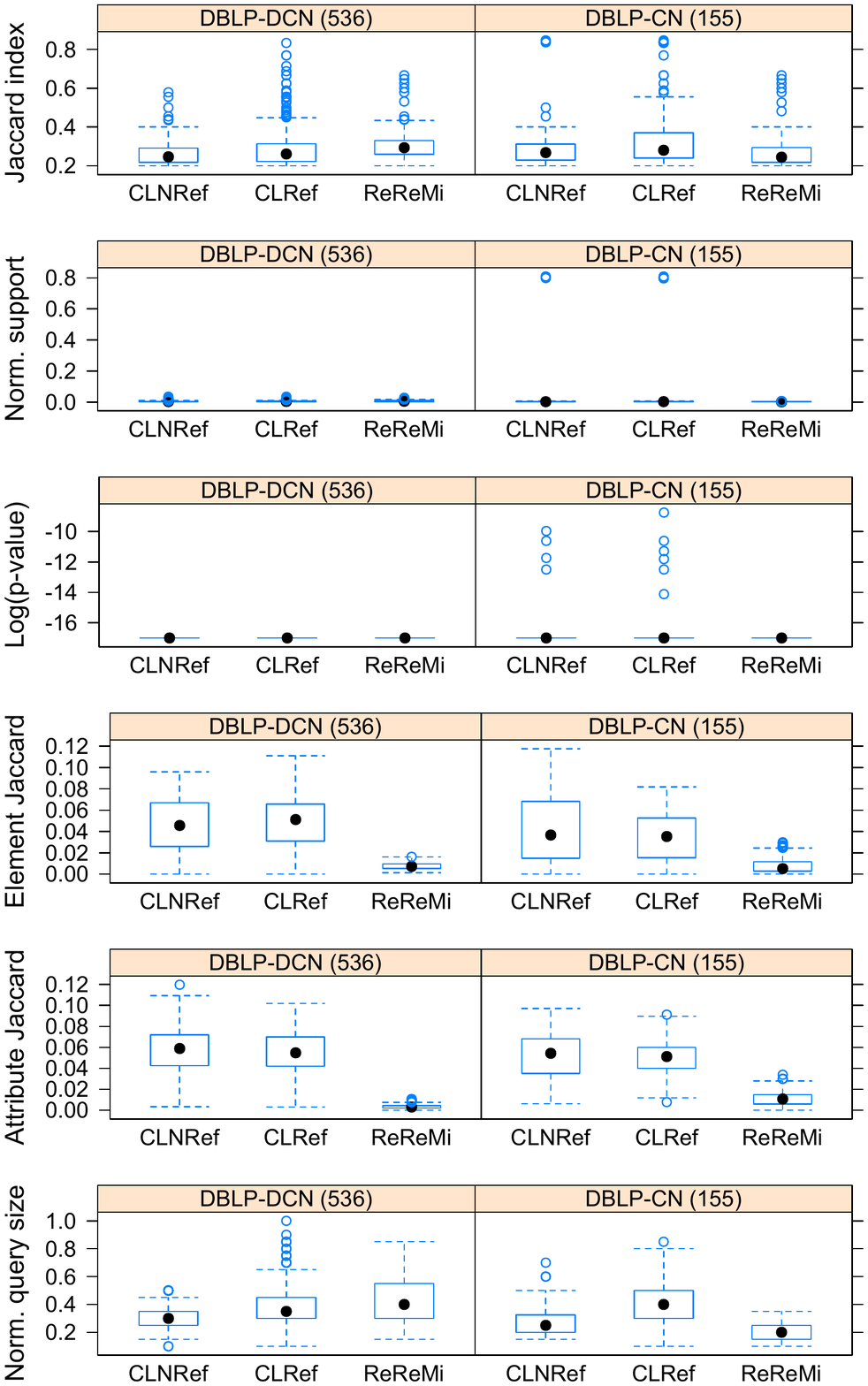}
\caption{Boxplots comparing redescriptions, produced with our framework (CLNref, CLRef) and the ReReMi algorithm (ReReMi) on the DBLP dataset. Sets contain $536$ redescriptions created by using all defined logical operators and $155$ redescriptions when only conjunction and negation operators are used to construct redescription queries.}
   \label{fig:RRDBLP}
\end{figure}

\begin{figure}[ht]
    \centering

  \includegraphics[width=0.5\textwidth]{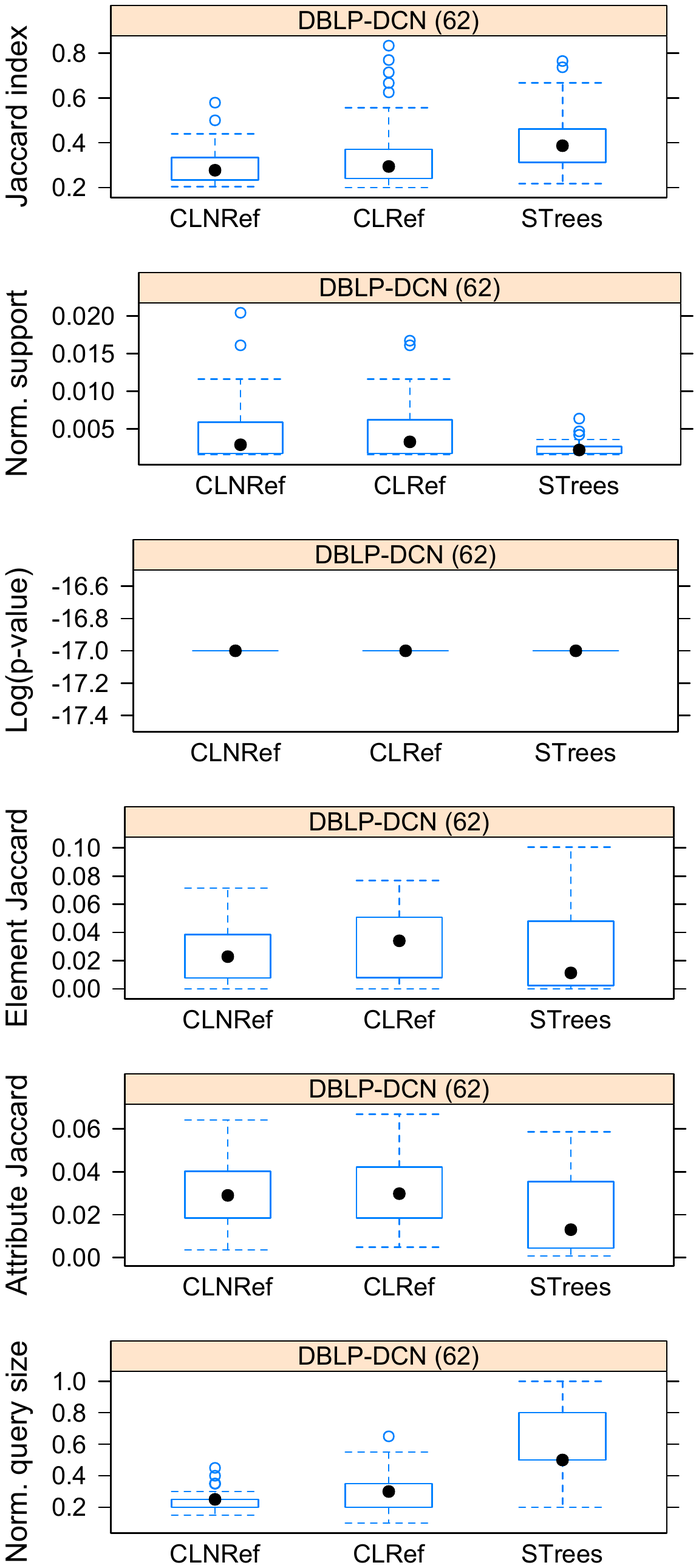}
\caption{Boxplots comparing redescriptions, produced with our framework (CLNref, CLRef) and the Split trees algorithm (STrees) on the DBLP dataset. The set contains $62$ redescriptions created by using all defined logical operators.}
   \label{fig:STDBLP}
\end{figure}

\noindent The DBLP dataset is very sparse and all redescription mining algorithms we tested only returned a very small number of highly accurate redescriptions. Half of the redescription mining runs  we performed with different algorithms returned to small number of redescriptions to perform a statistical analysis. On this dataset, we can compare quality measure distributions of redescriptions produced by our framework only with the ReReMi algorithm (Figure \ref{fig:RRDBLP}), and with the Split trees algorithm when all operators are used to construct redescription queries. (Figure \ref{fig:STDBLP}). 

\noindent CRM-GRS tends to produce redescriptions with smaller query size than the ReReMi and the Split trees algorithms when all the operators are allowed. The redescriptions contained in the reduced set produced by our framework tend to have higher support than those produced by the Split trees algorithm. The distribution analysis on sets created by using only conjunction and negation logical operators can be performed only against the ReReMi algorithm due to small number of redescriptions produced by the other approaches. In this case, CRM-GRS tends to produce more accurate redescriptions (significant at the significance level of $0.01$ when the conjunctive refinement is used and at the significance level of $0.05$ when conjunctive refinement is not used). In both cases, our framework produces redescriptions that tend to have larger support (significant with the level of $0.01$). There is a more pronounced difference between the Split trees algorithm and CRM-GRS when all the operators are allowed. In this case, the Split trees algorithm has higher median in distribution of redescription accuracy.

The Layered trees approach produced $7$ redescriptions using all operators, with accuracy $0.85,0.81,0.71,0.73,0.23,0.23,0.2$ describing $10$ to $48$ authors. It produced $3$ redescriptions using only conjunction and negation operators. The produced redescriptions had the accuracy $0.23,0.22,0.2$ and the support $45$ to $48$ authors. The Split trees algorithm produced only one redescription with accuracy $0.33$ and support $13$ using only conjunction and negation operators.

The most accurate redescriptions produced by each algorithm and a short discussion can be seen in Section S2.5.2 (Online Resource 1). 

\subsubsection{Comparison on the Country dataset}
\noindent Comparisons on the Country dataset are preformed only with the ReReMi algorithm since it is the only algorithm, besides CLUS-RM, that can work on datasets containing missing values. Techniques for value imputation must be used before other approaches can be applied. Using these techniques introduces errors in the descriptions and violates a property of descriptions being valid for each element in redescription support. Because of that, we chose not to pursue this line of research.

\begin{figure}[ht]
    \centering

  \includegraphics[width=0.5\textwidth]{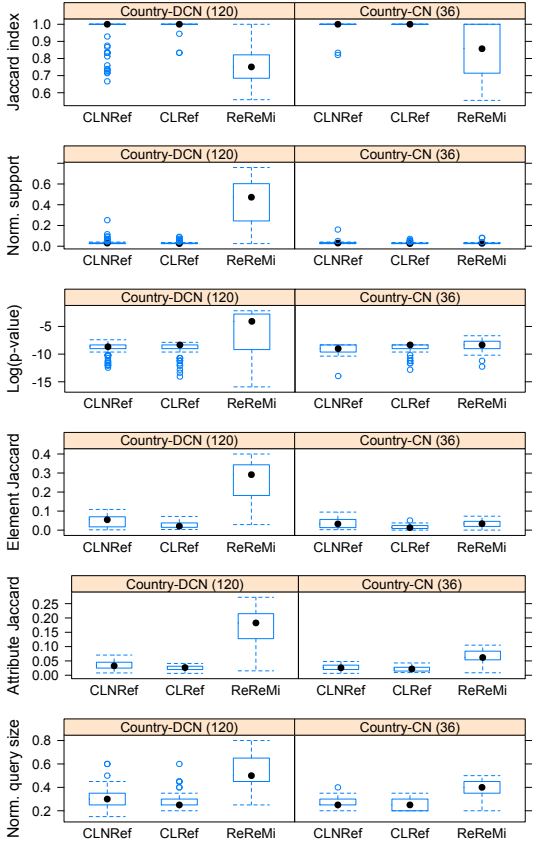}
\caption{Boxplots comparing redescriptions, produced with our framework (CLNref, CLRef) and the ReReMi algorithm (ReReMi) on the Country dataset. Sets contain $120$ redescriptions created by using all defined logical operators and $36$ redescriptions when only conjunction and negation operators are used to construct redescription queries. Redescription accuracy is evaluated by using query non - missing Jaccard index.}

   \label{fig:RRTrade}
\end{figure}

\begin{figure}[ht]
    \centering

  \includegraphics[width=0.5\textwidth]{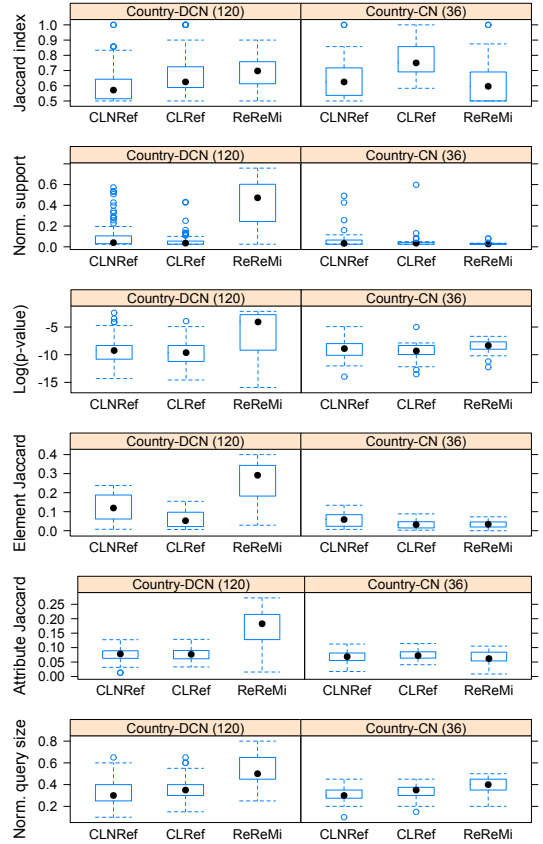}
\caption{Boxplots comparing redescriptions, produced with our framework (CLNref, CLRef) and the ReReMi algorithm (ReReMi) on the Country dataset. Sets contain $120$ redescriptions created by using all defined logical operators and $36$ redescriptions when only conjunction and negation operators are used to construct redescription queries. Redescription accuracy is evaluated by using pessimistic Jaccard index.}
   \label{fig:RRTradePess}
\end{figure}

Since our framework optimizes the query non-missing Jaccard index and the ReReMi optimizes pessimistic Jaccard index, we decided to make comparisons using both measures (Figure \ref{fig:RRTrade} and Figure \ref{fig:RRTradePess}). We extract two sets with CRM-GRS, for each we use different Jaccard index as one of the quality criteria. Redescriptions produced by the ReReMi remain unchanged but we compute the query non-missing Jaccard for each redescription which causes redescription accuracy to rise. Optimizing pessimistic Jaccard seems like the best option for comparisons since then the query non-missing Jaccard index necessarily increases and the redescription support is preserved.

Results from Table \ref{table1} show that CRM-GRS produces redescription set that tends to contain more accurate redescriptions when conjunction refinement procedure is used. The result is significant at the significance level $0.01$. However, it failed to produce such set using all operators when pessimistic Jaccard index is used to evaluate redescription accuracy (redescription set produced by ReReMi has higher median in accuracy). Although, CRM-GRS produced a few redescriptions with higher accuracy than those produced by the ReReMi. When query non-missing Jaccard index is used as accuracy evaluation criteria, CRM-GRS tends to create more accurate redescriptions than the ReReMi (statistically significant at the significance level $0.01$). When using only conjunction logical operator, the ReReMi tends to produce redescriptions with smaller support compared to CRM-GRS if conjunctive refinement procedure is not used. 

Analysis of element and attribute coverage is provided in Section S2.5.3 (Online Resource 1).

The ReReMi algorithm found $2$ redescriptions with $J_{pess}=1.0$ while CRM-GRS created redescription set containing $4$ redescriptions with $J_{pess}=1.0$ when only conjunction operators are allowed and $5$ redescriptions when all operators are allowed.

The analysis of comparative redescription examples produced by CRM-GRS and the ReReMi algorithm can be seen in Section S2.5.3 (Online Resource 1).

The ReReMi produced $14$ redescriptions with $J_{qnm}=1.0$ using only conjunction operators while redescription sets constructed by CRM-GRS contain $34$ out of $36$ redescriptions with $J_{qnm}=1.0$ without using conjunctive refinement and $36$ out of $36$ redescriptions with $J_{qnm}=1.0$ with the use of conjunctive refinement procedure. When all logical operators were used to create redescriptions, the ReReMi creates large number of disjunction based redescriptions, many of which are quite complex.

The difference in support size of redescriptions produced by CRM-GRS compared to those produced by the ReReMi algorithm, visible in Figures \ref{fig:RRTrade} and \ref{fig:RRTradePess} when all operators are used is in part the consequence of CRM-GRS using high weight on element diversity but is also connected to different logic in using the disjunction operator. CRM-GRS allows improving Jaccard index, by using disjunctions, only for redescriptions satisfying a predefined accuracy threshold. Highly overlapping subsets of instances are thus complemented with subsets that are highly overlapping with one of the already existing subset of instances. Because of this, our framework eliminates descriptions of unrelated subsets of instances that occasionally occur in ReReMi's descriptions as a result of using disjunction operator (discussed in \citep{GalbrunPhd}).

\section{Conclusions}
\label{concl}

\noindent We have presented a redescription mining framework CRM-GRS which integrates the generalized redescription set construction procedure with the CLUS-RM algorithm \citep{Mihelcic15,Mihelcic15LNAI}. 

The main contribution of this work is the generalized redescription set construction procedure that allows creating multiple redescription sets of reduced size with different properties defined by the user. These properties are influenced by the user through importance weights on different redescription criteria. Use of the scalarization technique developed in multi - objective optimization guarantees that, at each step, one non-dominated redescription is added to the redescription set under construction. The generalized redescription set construction procedure has lower worst time complexity than existing redescription mining algorithms so it may be preferred choice over the multiple runs of these algorithms. The procedure allows creating sets of different size with different redescription properties. These features generally lack in current redescription mining approaches, where users are forced to experiment with individual algorithm parameters in order to obtain desirable set of redescriptions. Finally, the procedure allows using ensembles of redescription mining algorithms to create reduced sets with superior properties compared to those produced by individual algorithms. 

The second contribution is related to increasing overall redescription accuracy. Here, we build upon our previous work on CLUS-RM algorithm and provide new - conjunctive refinement procedure, that significantly enlarges and improves the accuracy of redescriptions in the baseline redescription set by combining candidate redescriptions during the generation process. This procedure can be easily applied in the context of majority of other redescription mining algorithms, thus we consider it as a generally useful contribution to the field of redescription mining. 

Finally, we motivate the use of query non-missing Jaccard index, introduced in \citep{Mihelcic15LNAI}, when data contains missing values. We show that using pessimistic Jaccard index eliminates some potentially useful, high quality redescriptions obtainable by using query non-missing Jaccard index. To further increase the possibilities of redescription mining algorithms, we introduce the redescription variability index that allows extracting stable redescriptions in the context of missing data, by combining the upper and lower bound on estimates of Jaccard index.

The evaluation of our framework with $3$ different state of the art algorithms on $3$ different real-world datasets shows that our framework significantly outperforms other approaches in redescription accuracy in majority of cases. In particular in settings when only conjunction and negation operators are used in redescriptions, which is the preferred setting from the point of understandability. 
In general, CRM-GRS produces more understandable redescriptions (due to smaller query size and extensive use of conjunction operator), it is more flexible and in majority of comparisons more accurate approach to mine redescriptions from datasets. Moreover, we demonstrated that it complements existing approaches in the discovered redescriptions and solves several problems of existing approaches (mainly the problem of support drift and redescriptions connecting unrelated parts of element space by using disjunctions). The framework is easily extendible with new redescription criteria and allows combining results of different redescription mining algorithms to create reduced sets with superior properties with respect to different redescription quality criteria.

\subsection*{Acknowledgement}
{\par \vspace{1mm}\noindent \footnotesize \linespread{0.4}\selectfont The authors would like to acknowledge the European Commission's support through the  MAESTRA project (Gr. no. 612944), the MULTIPLEX project (Gr.no. 317532), and support of the Croatian Science Foundation (Pr. no. 9623: Machine Learning Algorithms for Insightful Analysis of Complex Data Structures).\par}

\section*{References}
\bibliography{mybibfile}

\end{document}